\documentclass{article}

    \usepackage[nonatbib,preprint]{neurips_2025}

\usepackage[square,numbers]{natbib}

\usepackage[utf8]{inputenc} 
\usepackage[T1]{fontenc}    
\usepackage{hyperref}       
\usepackage{url}            
\usepackage{booktabs}       
\usepackage{amsfonts}       
\usepackage{nicefrac}       
\usepackage{microtype}      
\usepackage[dvipsnames]{xcolor}         

\usepackage{amsmath}
\usepackage{amsthm}
\usepackage{graphicx}
\usepackage{wrapfig}
\usepackage{listings}
\usepackage{tikz}
\usepackage{subcaption}
\usepackage{etoolbox}
\usetikzlibrary{arrows.meta, positioning, calc, chains, shapes, shadows, patterns}
\usepackage{pgfplots}
\pgfplotsset{compat=1.17}
\usepgfplotslibrary{groupplots}
\usepackage{pgf-pie}
\usepackage{tikzviolinplots}

\newcommand{\kiviatgrid}[4]{
  \foreach \i in {1,...,#3} {
    \draw[gray!60] #1 -- ++({90+360/#3*(\i-1)}:#2);
  }
  \foreach \l [evaluate={\r=\l*#2/#4;}] in {1,...,#4} {
    \path #1 coordinate (center);
    \foreach \i in {1,...,#3} {
      \path (center) ++({90+360/#3*(\i-1)}:\r) coordinate (P\l-\i);
    }
    \draw[gray!30] (P\l-1)
      \foreach \i in {2,...,#3} {
        -- (P\l-\i)
      } -- cycle;
  }
}

\newtheorem{theorem}{Theorem}
\newtheorem{definition}{Definition}

\newcommand\independent{\protect\mathpalette{\protect\independenT}{\perp}}
\def\independenT#1#2{\mathrel{\rlap{$#1#2$}\mkern2mu{#1#2}}}

\title{Causal Cartographer: From Mapping to Reasoning Over Counterfactual Worlds}

\author{
    Gaël Gendron$^1$\thanks{Corresponding author. Email: \texttt{ggen187@aucklanduni.ac.nz}} \quad Jože M Rožanec$^{2,3}$ \quad Michael Witbrock$^1$ \quad Gillian Dobbie$^1$ \\
    $^1$NAOInstitute, University of Auckland \\ 
    $^2$Jožef Stefan Institute \\
    $^3$University of Twente \\
}

\newcommand{\printlistNodes}[1]{%
\begingroup
\renewcommand{\do}[1]{%
\noindent\tikz\node[extraattr, inner sep=2pt]{##1};\quad%
}%
\docsvlist{#1}%
\endgroup
}

\newcommand{\ParameterNode}[5]{%
\tikzset{%
    paramnode/.style={
    rectangle, rounded corners,
    draw=black, very thick,
    font=\sffamily\scriptsize,
    fill=blue!20,
    drop shadow,
    align=center,
    inner sep=5pt,
    },
    extraattr/.style={
    draw, 
    rounded corners, 
    fill=orange!20, 
    inner sep=2pt}
}
\node[paramnode] (param) {%
\begin{tabular}{@{}c@{}}
\textbf{#1} \\[1mm]
(\textit{#3}: #4) \\[1mm]
\tikz\node[text width=0.7\linewidth]{#2}; \\[1mm]
\parbox[t]{0.5\linewidth}{\printlistNodes{#5}} 
\end{tabular}
};
}

\begin{document}

\maketitle

\begin{abstract}

Causal world models are systems that can answer counterfactual questions about an environment of interest, i.e. predict how it would have evolved if an arbitrary subset of events had been realized differently. It requires understanding the underlying causes behind chains of events and conducting causal inference for arbitrary unseen distributions.
So far, this task eludes foundation models, notably large language models (LLMs), which do not have demonstrated causal reasoning capabilities beyond the memorization of existing causal relationships.
Furthermore, evaluating counterfactuals in real-world applications is challenging since only the factual world is observed, limiting evaluation to synthetic datasets. 
We address these problems by explicitly extracting and modeling causal relationships and propose the Causal Cartographer framework. First, we introduce a graph retrieval-augmented generation agent tasked to retrieve causal relationships from data.
This approach allows us to construct a large network of real-world causal relationships that can serve as a repository of causal knowledge and build real-world counterfactuals.
In addition, we create a counterfactual reasoning agent constrained by causal relationships to perform reliable step-by-step causal inference. 
We show that our approach can extract causal knowledge and improve the robustness of LLMs for causal reasoning tasks while reducing inference costs and spurious correlations.
  
\end{abstract}

\section{Introduction}
\label{sec:intro}

Learning to infer causal relationships and making causal predictions about the world is an important task for general reasoning systems \citep{DBLP:journals/corr/abs-2011-15091,DBLP:journals/pieee/ScholkopfLBKKGB21}. In particular, predicting how an environment would have evolved under a different policy (i.e. counterfactual questions) is a challenging and crucial question when evaluating how an artificial system understands the world. While an agent with low causal knowledge can make predictions about observed distributions, generalizing to arbitrary distributions and counterfactuals requires one to learn a causal world model \citep{DBLP:books/acm/22/BareinboimCII22,DBLP:conf/iclr/RichensE24}. Studies on large language models (LLMs) have shown that they do not perform robust causal discovery or causal inference \citep{DBLP:journals/tmlr/ZecevicWDK23,DBLP:conf/nips/JinCLGKLBAKSS23,DBLP:conf/iclr/Jin0LPSMDS24,DBLP:conf/emnlp/ChenXWZZZL24,DBLP:journals/corr/abs-2402-01207,DBLP:conf/emnlp/JoshiSW024} and fail to generalize to unseen distributions \citep{DBLP:conf/naacl/WuQRA0WKAK24,DBLP:conf/ijcai/Gendron0WD24,DBLP:conf/emnlp/GendronNPWD24,DBLP:conf/iclr/BerglundTKBSKE24}. This is a challenging task as causal knowledge is notoriously hard to collect \citep{rubin1974estimating,pearl2009causality} and counterfactual data is generally not available as only the factual world is observed \citep{holland1986statistics}.

We tackle these problems by explicitly modeling causal relationships and enhancing LLM agents with a causal reasoning framework. We introduce the \textbf{Causal Cartographer}:
a twofold system composed of a \textit{graph retrieval-augmented generation} (graphRAG \citep{DBLP:conf/nips/LewisPPPKGKLYR020,DBLP:journals/corr/abs-2404-16130,DBLP:journals/corr/abs-2408-08921}) \textit{agent} tasked to retrieve causal relationships from real-world news articles and a \textit{counterfactual reasoning agent} performing reliable and efficient step-by-step causal inference while respecting causal relationships.
We take advantage of the first component to build CausalWorld, a causal network that maps causal knowledge of the world, which we use to \textit{provably} access counterfactual knowledge and build real-world causal questions for our second reasoning agent. We use the extracted knowledge to guide our causal reasoning agent on real-world counterfactual reasoning tasks. We show that our proposed method allows for the extraction of causal knowledge from the data and the evaluation of agents in real-world counterfactual situations.
We summarize our contributions as follows:
\begin{itemize}
    \item We introduce a causal extraction method for unstructured natural language data and a graph retrieval-augmented generation (Graph-RAG) agent;
    \item We use our method to build CausalWorld, a network of causal relationships from real-world news events published in 2020, containing 975 variables and 1337 causal relationships;
    \item We use this network as a repository of causal knowledge to evaluate large language models on counterfactual reasoning with real-world natural language data;
    \item We introduce the theoretical notion of causal blankets and prove that, assuming identifiability of the blanket, our method allows estimating true counterfactuals;
    \item We introduce an auxiliary step-by-step causal inference agent that can perform counterfactual reasoning respecting causal constraints;
    \item We show that causal conditioning achieves competitive performance for counterfactual reasoning while greatly reducing the LLM's context window and output length, decreasing the inference cost up to 70\%.
    
\end{itemize}

Our code is available at: \url{https://github.com/ggendro/causal-cartographer}.

\section{Background}
\label{sec:background}

\paragraph{Structural Causal Models}

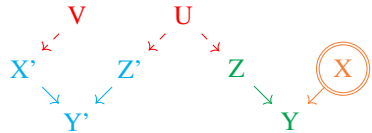
\begin{wrapfigure}{r}{0.5\linewidth}
    \centering
    \begin{tikzpicture}[
            node distance=1cm,
            factual/.style={Cyan},
            intervention/.style={Orange},
            counterfactual/.style={Green},
            latent/.style={red}
        ]
    
        \node[factual] (x) {X'};
        \node[latent] (v) [above right of=x] {V};
        \node[factual] (y) [below right of=x] {Y'};
        \node[factual] (z) [above right of=y] {Z'};
        \node[latent] (u) [above right of=z] {U};
        \node[counterfactual] (z2) [below right of=u] {Z};
        \node[counterfactual] (y2) [below right of=z2] {Y};
        \node[draw,circle,double,intervention] (x2) [above right of=y2] {X};
    
        \draw[->,factual] (x) -- (y);
        \draw[->,factual] (z) -- (y);
        \draw[->,dashed,latent] (v) -- (x);
        \draw[->,dashed,latent] (u) -- (z);
        \draw[->,dashed,latent] (u) -- (z2);
        \draw[->,intervention] (x2) -- (y2);
        \draw[->,counterfactual] (z2) -- (y2);
    \end{tikzpicture}
    \caption{Counterfactual twin graph for three endogenous variables $\{\textcolor{Cyan}{X}, \textcolor{Cyan}{Y}, \textcolor{Cyan}{Z}\}$ and two exogenous variables $\{\textcolor{red}{U},\textcolor{red}{V}\}$, under an intervention over $\textcolor{Orange}{X}$. A counterfactual world is identical to the factual world, except for the intervention and its effects. Exogenous variables are shared by both worlds, connecting the two graphs. $\textcolor{Orange}{X}$ is being intervened upon and so has no incoming edges. The counterfactual value of $\textcolor{Green}{Y}$ given by $P(\textcolor{Green}{Y} \mid \textcolor{Orange}{do(X)},\textcolor{Cyan}{X'},\textcolor{Cyan}{Y'})$ is obtained by estimating the value of $\textcolor{red}{U}$ from the factual observations, then deducing $\textcolor{Green}{Z}$ from $\textcolor{red}{U}$ (left unchanged) and $\textcolor{Green}{Y}$ from $\textcolor{Green}{Z}$ and $\textcolor{Orange}{X}$. }
    \label{fig:counterfactual_twin_graph}
\end{wrapfigure}

Recovering causal relationships in the real world is a long-standing problem of science \citep{pearl2009causality}. Structural Causal Models (SCMs) \citep{pearl2009causality, pearl2014deductive, DBLP:conf/nips/XiaLBB21} are graphical models that allow representing causal knowledge and performing causal inference. 
SCMs can be represented as Directed Acyclic Graphs (DAGs) where nodes are causal variables and edges are causal relationships. Causal variables can be distinguished between \textit{endogenous} (observed) and \textit{exogenous} (unobserved) variables. Exogenous variables $\mathbf{U}$ do not have parents and are represented by a probability distribution. Endogenous variables $\mathbf{V}$ are defined by functions that links them to their causal parents, i.e. $V_i \leftarrow f_{V_i}(\mathbf{pa}(V_i))$. SCMs are Markovian processes, which implies that if the full set of exogenous variables is known, then the values of every endogenous variable can be deterministically computed by iteratively inferring children values from their parents \citep{DBLP:journals/corr/abs-2302-00293}.
While very powerful tools to estimate causal effects, SCMs typically cannot be fully retrieved as causal information is challenging to obtain.

\paragraph{Counterfactual Inference}

Even without an SCM, some causal questions can be answered. Causal inference queries can typically be divided into three categories: \textit{observations}, \textit{interventions} and \textit{counterfactuals}, forming \textit{Pearl's Causal Hierarchy} \citep{DBLP:books/acm/22/BareinboimCII22}. Each category is harder to answer than the previous one, as it typically requires a better understanding of the causal relationships underlying the system of interest \citep{DBLP:books/acm/22/BareinboimCII22}. While multiple Markov-equivalent causal graphs can account for the same set of observations, answering interventions and counterfactuals requires one to know the true local causal structure \citep{DBLP:books/acm/22/BareinboimCII22, DBLP:conf/iclr/RichensE24}. Moreover, the result of a counterfactual is generally not accessible in real-world scenarios. This is the \textit{fundamental problem of causal inference}: only the factual world is observed~\citep{holland1986statistics}.

Observational queries are represented as conditional probabilities $P(Y \mid X)$. Interventions are represented as $P(Y \mid do(X))$ using the \textit{do-operator} \citep{pearl2009causality, DBLP:conf/uai/Pearl12}. An intervention alters the causal structure by forcing the value of the variable $X$, regardless of its prior probability, effectively cutting the parents of $X$. An intuitive example is a randomized control trial where the probability of obtaining treatment is randomized to ensure that it cannot correlate with other factors. Counterfactuals correspond to ``what if?'' questions, asking how would a world evolve under an intervention, given the outcome in the factual world. They can be represented with the following equation: 

\begin{equation}
    \label{eq:counterfactual-exogenous}
    P(Y \mid do(X),X',Y') = \sum_{U \in \mathbf{U}} P(Y \mid do(X),U)P(U \mid X',Y')  
\end{equation}

$X'$ and $Y'$ represent the factual observations. $X$ and $Y$ correspond to the variables in the counterfactual world. $\mathbf{U}$ is the set of exogenous variables shared between the two worlds. Figure \ref{fig:counterfactual_twin_graph} shows a graphical expression with twin graphs corresponding to the factual and counterfactual worlds.

\section{Graph Retrieval-Augmented Causal Extraction}
\label{sec:causal_extraction}

This work introduces a method for causal extraction and reasoning based on LLM agents. This section focuses on the \textit{Causal Cartographer Extraction agent} (CTG-Extract). We use it to recover existing causal knowledge about the world from news sources. The resulting causal network, insights we gather from it, and its application to counterfactuals are discussed in Sections \ref{sec:causal_world} and \ref{sec:counterfactual-inference}.

\subsection{CTG-Extract}

\begin{figure}
    \begin{subfigure}{0.48\textwidth}
        \centering
        \includegraphics[width=0.9\linewidth]{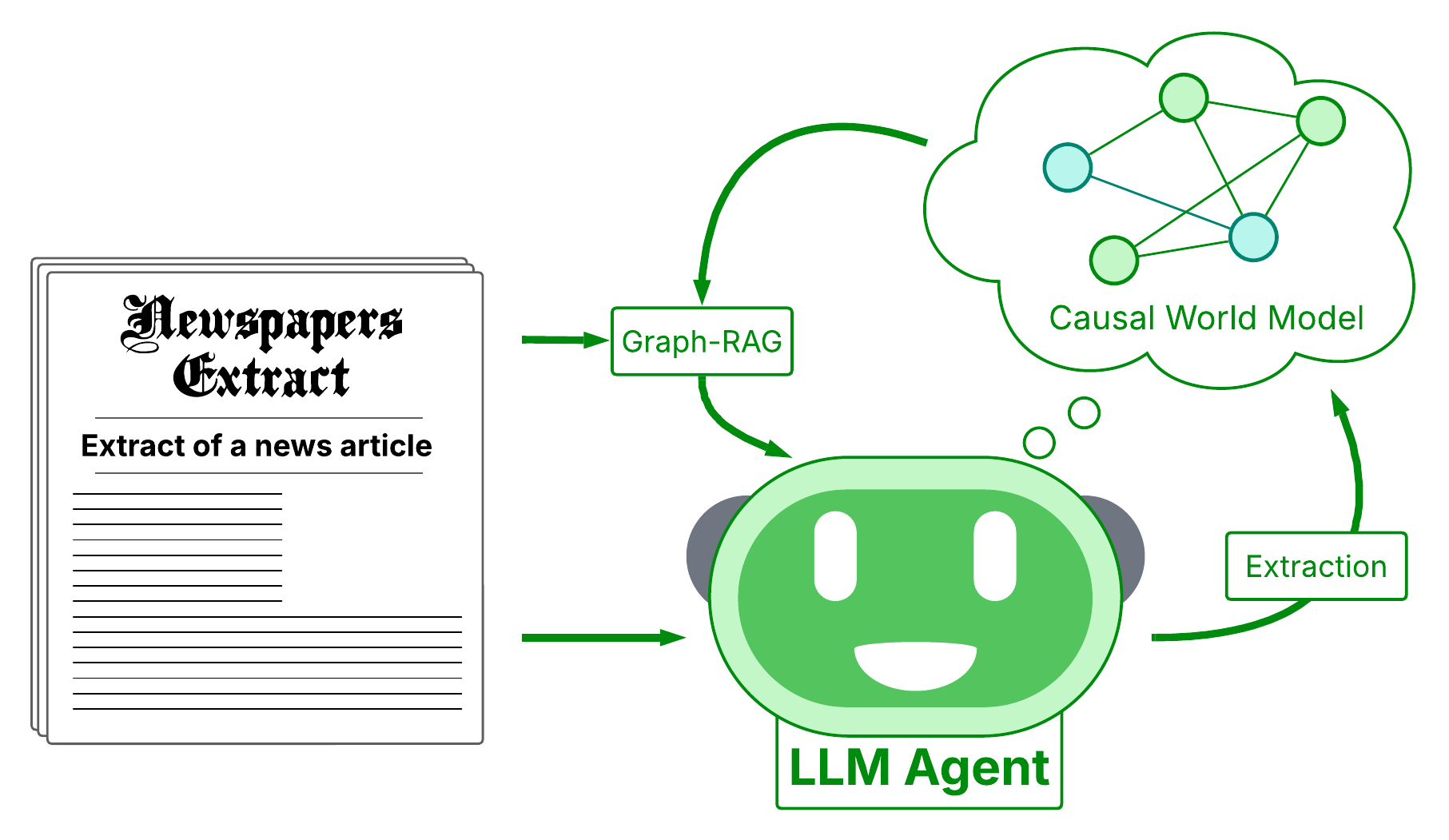}
        \caption{Causal extraction process.  }
        \label{fig:causal-extraction}
    \end{subfigure}
    \hfill
    \begin{subfigure}{0.48\linewidth}
        \centering
        \begin{tikzpicture}
        \ParameterNode
        {India Oil Diversification Strategy}
        {India's strategic initiative to diversify its oil supply sources away from the traditional reliance on the Middle East.}
        {boolean}
        {[True, False]}
        {world\_188, world\_226, world\_363}
        \end{tikzpicture}
        \caption{Example of extracted node.}
        \label{fig:node-attrs-example}
    \end{subfigure}
    \caption{Overview of the \textit{Causal Cartographer Extraction agent} (CTG-Extract). (a) Illustration of the extraction pipeline. The title, content, and metadata from a news source are provided to the agent context. It then generates causal variables and their relationships and adds them to the causal world model. A graphRAG system further compares article data and agent node suggestions and adds to the agent context the nodes and causal relationships that seem related. (b) Example of a node extracted with the method. Its name, description, type, possible values, and the worlds it appears in are shown. }
    \label{fig:causal-extraction-main}
\end{figure}

We extract causal variables and causal relationships from news sources. Figure \ref{fig:causal-extraction} illustrates the causal extraction process. While causal structure discovery methods attempt to infer the causal structure from data, causal extraction tasks recover \textit{stated} causal relationships in text data \citep{DBLP:journals/corr/abs-2302-00293,DBLP:journals/kais/YangHP22}. Large language models have shown limited performance in discovering causal relationships beyond domain knowledge \citep{DBLP:conf/emnlp/JoshiSW024,DBLP:journals/corr/abs-2402-01207,DBLP:journals/tmlr/ZecevicWDK23,DBLP:conf/iclr/Jin0LPSMDS24}. Acknowledging this limitation, we rely on human sources and only use the LLM agent to perform causal extraction. Causal information can either be \textit{explicitly} or \textit{implicitly} stated in the text, within a single sentence or across multiple sentences of paragraphs. Previous methods based on pattern-matching or statistical modeling do not capture the latter well~\citep{DBLP:journals/kais/YangHP22}. Attention in LLMs allows them to aggregate information across a document, mitigating this problem.

\paragraph{Agent Description}
We construct an LLM agent following the ReAct framework \citep{DBLP:conf/iclr/YaoZYDSN023} and using the Smolagents library \footnote{https://huggingface.co/docs/smolagents/}. We prompt the LLM agent to conduct step-by-step causal extraction as a sequence of \textit{thoughts}, \textit{code} and \textit{observations}. A thought step is a draft of the step to execute while the code step is its implementation. The agent is provided with a Python interpreter. The standard output and errors of the interpreter are returned to the agent in the observation step. The agent then adapts its next step based on this feedback. It is tasked to execute the following plan:

\begin{enumerate}
    \item Define the causal variables existing in the input text;
    \item Define the \textit{confounders}, i.e. causal variables that are not observed or mentioned in the text but have a direct effect on the observed variables;
    \item Match the new variables with ones existing in the causal graph if possible (with graphRAG);
    \item Define the causal relationships between the variables based on the text, relationships already existing in the graph should not be created;
    \item Add the new causal variables and relationships to the graph.
\end{enumerate}

Every requirement can be executed in a single thought-code-observation cycle but the agent can perform more steps if it did not immediately succeed. The prompts are provided in Appendix \ref{app:prompts}. 
Each variable contains the following attributes: \texttt{name} and \texttt{description}, \texttt{type} and \texttt{values} describe the domain of the variable, \texttt{current\_value} is the current instantiation of the variable, \texttt{contextual\_information} provides additional context to the variable instantiation and \texttt{supporting\_text\_snippets} is the text extract justifying the response.
Each relationship contains \texttt{cause} and \texttt{effect} variables, and similar \texttt{description} and \texttt{contextual\_information} attributes.
Their implementation is provided in Appendix~\ref{app:prompts}. An example is also provided in Figure~\ref{fig:node-attrs-example}.

\paragraph{Grounding in Worlds}
To later perform counterfactual matching (in Section \ref{sec:counterfactual-inference}), we save multiple \textit{worlds} per node. Each document describes a world, i.e. an instantiation of the observed subgraph of the world graph. For instance, two nodes $A$ and $B$ can be described in several documents. The concept they represent (described by attributes \texttt{name}, \texttt{description}, \texttt{type} and \texttt{values}) is invariant to world changes. However, their instantiations (described by attributes \texttt{current\_value}, \texttt{contextual\_information}, and \texttt{supporting\_text\_snippets}) can differ from one world to the next. They are saved as coming from different worlds, as illustrated by Figure \ref{fig:node-attrs-example}.

\paragraph{NxGraphRAG}
In addition to the Python interpreter, we introduce a novel graph retrieval-augmented generation (graphRAG) method \citep{DBLP:conf/nips/LewisPPPKGKLYR020,DBLP:journals/corr/abs-2404-16130,DBLP:journals/corr/abs-2408-08921} that we equip the agent with.
The system, NxGraphRAG, generates an embedding of the input document and of each node of the causal graph using an auxiliary LLM and keeps them in a vector database. Before providing the document to the agent, NxGraphRAG is called and returns the top-$K$ nodes with the highest cosine similarity to the input document. It also traverses the graph through $P$ steps from each retrieved node and adds each neighbor node and edge to ensure that no relevant context is omitted. 
The embedding-based retrieval focuses on semantically relevant nodes while the traversal takes advantage of structural knowledge. After this initial call, the NxGaphRAG pipeline can be called again by the agent when creating new variables to verify that they do not already exist in the graph. An overview of the pipeline and additional details are provided in Appendix \ref{app:rag_details}.

\subsection{Data}
Our effort to build a causal graph from phenomena observed in the real world is part of a broader effort to understand the factors that influence the investment in and use of green energy \citep{rovzanec2023building,DBLP:journals/corr/abs-2410-06392}. 
Therefore, when building the dataset, we constrained ourselves to media news related to the \textit{oil price} topic, given research has shown that oil price fluctuations affect the consumption of renewable energy sources \cite{troster2018renewable}, and that positive shocks in oil prices reduce CO\textsubscript{2} emissions \cite{hu2022oil}.
We constructed a dataset of 500 media events from 2020 by extracting them from EventRegistry \cite{DBLP:conf/www/LebanFBG14}, a platform sourcing, processing, and clustering media news based on events reported over time. 
For each media event, they provide a summary and information regarding the piece of news considered the most representative. 
We exploited this information to build the causal graph described in Section \ref{sec:causal_world}.

\section{The CausalWorld Network}
\label{sec:causal_world}

\begin{figure}
    \centering
    \includegraphics[width=\linewidth,trim=0cm 1cm 0cm 1cm,clip]{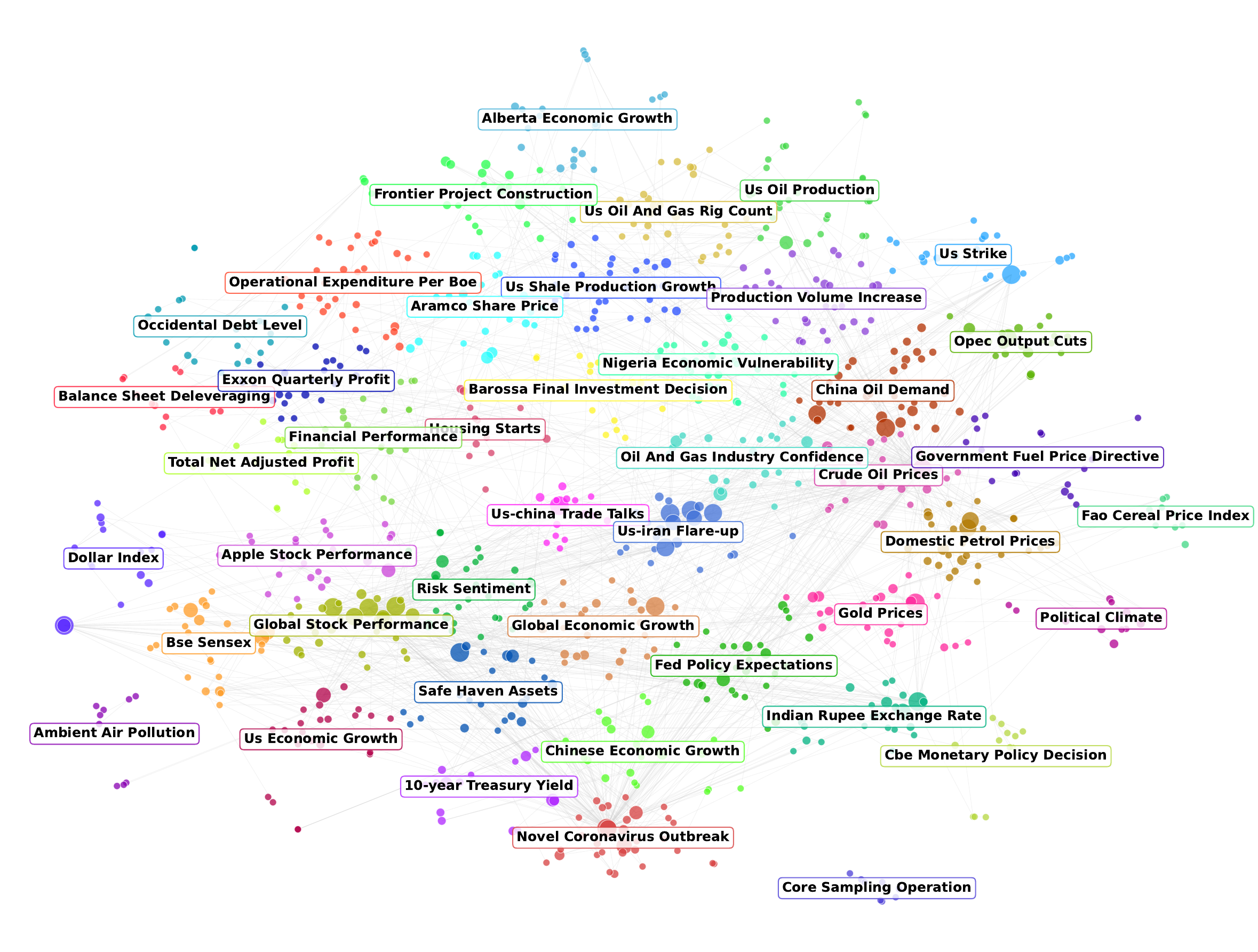}
    \caption{CausalWorld graph structure. Colors show the cluster in which the node belongs. Node size is based on the number of worlds a node appears in: the more a node appears, the bigger it is on the visualization. Labels correspond to the name of the most representative node of the cluster. }
    \label{fig:causal-world-graph-structure}
\end{figure}

The CausalWorld network is a causal graph generated using the methodology described in Section~\ref{sec:causal_extraction}. It contains 975 nodes and 1337 edges. In this section, we detail the findings obtained from the network. Figure \ref{fig:causal-world-graph-structure} shows a visualization of the network.
The resulting structure of CausalWorld is sparse, with a graph density of $\sim 0.001$. This is in accordance with the Sparse Mechanisms Shift (SMS) hypothesis, which states that a causal mechanism only sparsely affects other mechanisms~\citep{DBLP:journals/pieee/ScholkopfLBKKGB21}. Figure \ref{fig:causal-world-components} in Appendix \ref{app:causalworld_structure} further shows the distribution of strongly and weakly connected components, highlighting that the majority of the network is connected. Despite this structure, few feedback loops are observed. 
The network encompasses diverse knowledge and can be divided into 44 topic clusters like \texttt{Gold Prices} or \texttt{US Oil Production}, as shown in Figure \ref{fig:causal-world-graph-structure}.
Furthermore, 109 structural communities can be found, communicating with other communities with a small set of bridge nodes, e.g.: \texttt{Crude Oil Prices}, \texttt{Novel Coronavirus Outbreak}, \texttt{Global Economic Growth} and \texttt{US-China Trade Talks}. These nodes are key elements of the graph that enable the propagation of information from one community to the next.
More details are provided in Appendix~\ref{app:causalworld_structure}.

\subsection{The CausalWorld Network combines Many Overlapping Worlds}

The main difference between CausalWorld and other repositories of causal relationships is the presence of \textbf{worlds}, as described in Section~\ref{sec:causal_extraction}. Each world instantiates a variable by providing it with a value and context. It is the key to match worlds and perform counterfactuals (discussed in Section \ref{sec:counterfactual-inference}). However, worlds must not be isolated and communicate via nodes shared by multiple worlds. 37\% of the nodes in CausalWorld share two worlds or more, allowing information to be propagated across worlds. Since the majority of the nodes are linked by bridge nodes, most nodes can be used to compute counterfactuals.

\subsection{Causal Insights from Structure: ``Biofuel Demand Impacts Food Prices''}

Interesting findings that can be extracted from the network structure are causal paths established by the causal extraction process. 
We provide an example of a causal chain in Figure \ref{fig:causal-chain-ex}. 
We can observe causal relationships between nodes extracted from different documents, e.g. ``Palm oil Prices'' belongs to world 119 but ``Food Prices'' connects it to worlds 58, 68, and 70. It illustrates a major property of CausalWorld: it allows inference to be performed between long-range dependencies across multiple sources while maintaining information sparsity. 
CausalWorld also allows for cycles in the graph and can, thus, represent feedback loops, as illustrated in Figure \ref{fig:causal-cycle-ex}. More examples are provided in Appendix \ref{app:causal-paths}.

\begin{figure}
    \centering
    \begin{tikzpicture}[
        block/.style={
        rectangle, rounded corners,
        draw=black, very thick,
        font=\scriptsize,
        fill=blue!20,
        drop shadow, 
        minimum width=2cm, minimum height=1cm,
        align=center,
        inner sep=5pt,
        text width=2cm,
        },
        process/.style={
        rectangle, rounded corners,
        draw=black, very thick, dashed,
        font=\scriptsize,
        fill=red!20,
        drop shadow, 
        minimum width=2cm, minimum height=1cm,
        align=center,
        inner sep=5pt,
        text width=2cm,
        },
        causal arrow/.style={
        -{Stealth[length=4mm, width=3mm]},
        thick,
        shorten >=2pt
        },
        every node/.append style={on chain, node distance=1.5cm}
        ]
        \begin{scope}[start chain=going right]
            \node[block] (A) at (0,0) {Biofuel Demand};
            \node[process] (B) {Strong biofuel demand drives up palm oil prices.};
            \node[block] (C) {Palm Oil Prices};
            \node[process] (D) {Higher palm oil prices increase the vegetable oil price index.};
        \end{scope}
    
        \begin{scope}[start chain=going left, yshift=-1.5cm]
            \node[block] (E) at (0.815\linewidth,0) {FAO Vegetable Oil Price Index};
            \node[process] (F) {An increased vegetable oil price index raises overall food prices.};
            \node[block] (G) {Food Prices};
            \node (H) {...};
        \end{scope}
    
        \draw[->,causal arrow] (A) -- (B);
        \draw[->,causal arrow] (B) -- (C);
        \draw[->,causal arrow] (C) -- (D);
        \draw[->,causal arrow] (D) -- (E);
        \draw[->,causal arrow] (E) -- (F);
        \draw[->,causal arrow] (F) -- (G);
        \draw[->,causal arrow] (G) -- (H);
    \end{tikzpicture}
    \caption{Illustration of a direct causal path in the CausalWorld graph. Nodes are \textcolor{blue!60}{blue} boxes. Arrows represent causal dependencies. The description of the dependency is shown in dashed \textcolor{red!40}{red} boxes. Note that all nodes except for the root can have additional causal parents not shown in the chain and that the strength and function related to the causal relationships are not shown. }
    \label{fig:causal-chain-ex}
\end{figure}
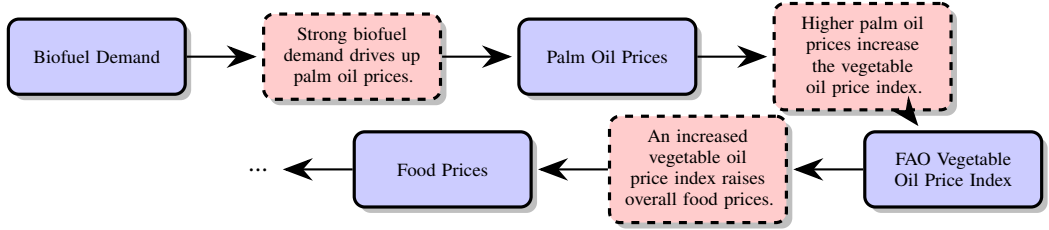

\begin{figure}
    \centering
    \begin{tikzpicture}[
        block/.style={
        rectangle, rounded corners,
        draw=black, very thick,
        font=\scriptsize,
        fill=blue!20,
        drop shadow, 
        minimum width=2cm, minimum height=1cm,
        align=center,
        inner sep=5pt,
        text width=2cm,
        },
        process/.style={
        rectangle, rounded corners,
        draw=black, very thick, dashed,
        font=\scriptsize,
        fill=red!20,
        drop shadow, 
        minimum width=2cm, minimum height=1cm,
        align=center,
        inner sep=5pt,
        text width=4cm,
        },
        causal arrow/.style={
        -{Stealth[length=4mm, width=3mm]},
        thick,
        shorten >=2pt
        },
        every node/.append style={on chain, node distance=2cm}
        ]
        
        \begin{scope}[start chain=going right]
            \node[block] (A) at (0,0) {Global Economic Growth};
            \node[process] (B) {Improved global economic growth leads to a higher oil demand forecast.};
            \node[block] (C) {Global Oil Demand Forecast};
        \end{scope}
        
        \begin{scope}[start chain=going left, yshift=-1.5cm]
            \node[process] (D) at (0.7\linewidth,0) {A lower forecast for global oil demand exerts downward pressure on crude oil prices.};
            \node[block] (E) {Crude Oil Prices};
            \node[process] (F) {Surging crude oil prices risk undermining global economic recovery by increasing costs and market uncertainty.};
        \end{scope}
    
        \draw[->,causal arrow] (A) -- (B);
        \draw[->,causal arrow] (B) -- (C);
        \draw[->,causal arrow] (C) -- (D);
        \draw[->,causal arrow] (D) -- (E);
        \draw[->,causal arrow] (E) -- (F);
        \draw[->,causal arrow] (F) -- (A);
    \end{tikzpicture}
    \caption{Illustration of a cycle in the CausalWorld graph. The legend is the same as in Figure \ref{fig:causal-chain-ex}. }
    \label{fig:causal-cycle-ex}
\end{figure}
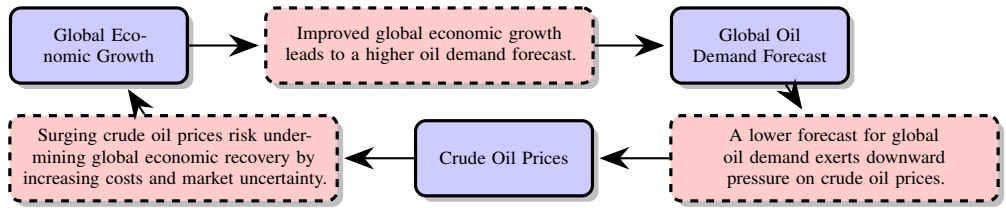

\section{Counterfactual Inference in the Real World}
\label{sec:counterfactual-inference}

\begin{figure}
    \centering
    \begin{subfigure}{0.56\linewidth}
        \includegraphics[width=\linewidth]{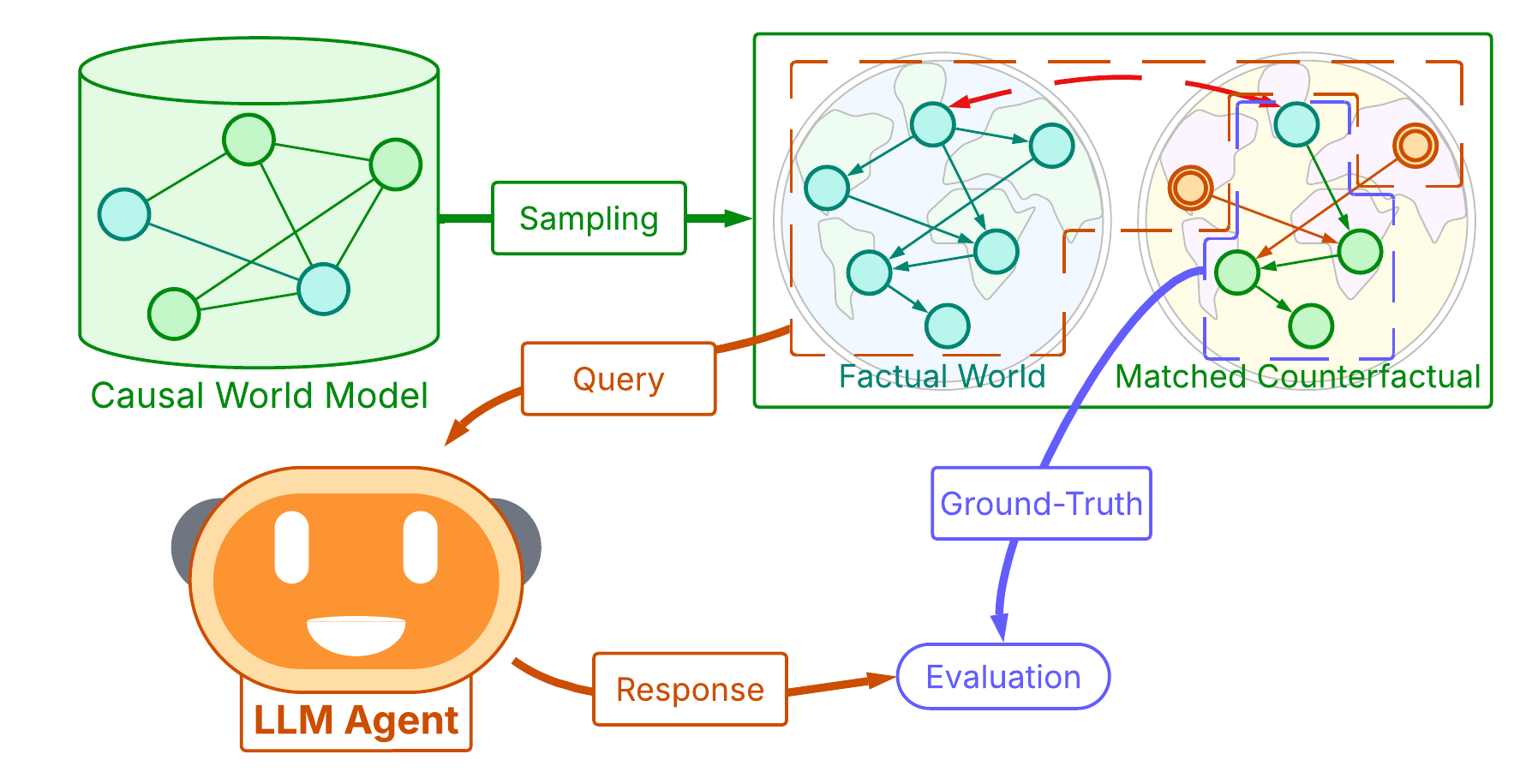}
        \caption{Overview of the counterfactual matching and evaluation. }
        \label{fig:counterfactual-matching}
    \end{subfigure}
    \hfill
    \begin{subfigure}{0.38\linewidth}
        \includegraphics[width=\linewidth]{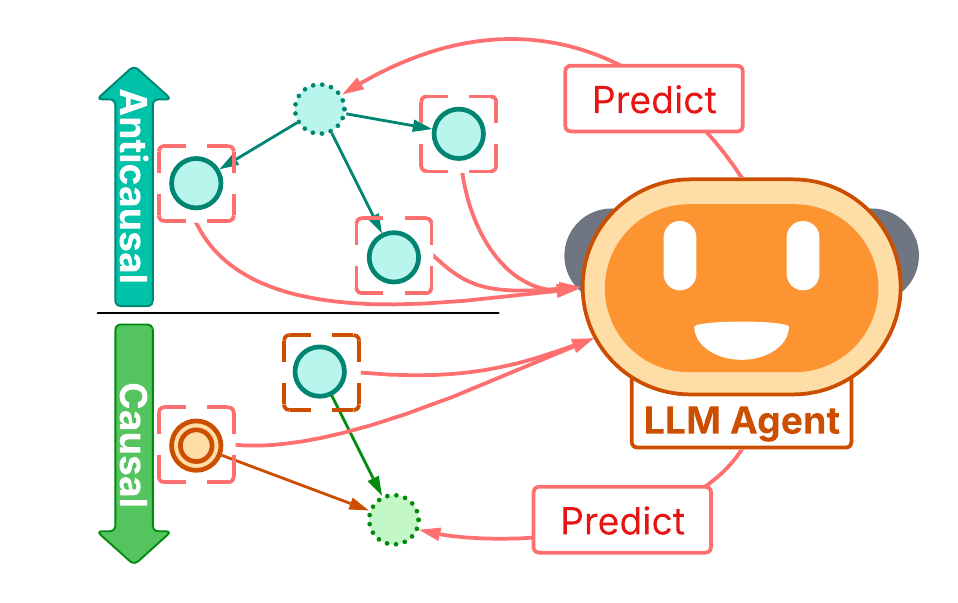}
        \caption{Details of the step-by-step inference.}
        \label{fig:step-by-step-cr}
    \end{subfigure}
    \caption{Overview of the \textit{Causal Cartographer Reasoning agent} (CTG-Reason) and the evaluation pipeline. (a) A query is sampled from the CausalWorld graph using counterfactual matching: values in the counterfactual world that match those in a factual world are abducted and replaced with information from the factual world. The agent performs counterfactual inference from the resulting graph. (b) The agent performs step-by-step reasoning by predicting the value of a child variable given its parents (causal direction), resp. predicting the value of a parent given its children (anticausal). }
    \label{fig:counterfactual-reasoning-agent}
\end{figure}

We now investigate how to take advantage of the CausalWorld graph to build a counterfactual reasoning evaluation dataset (CausalWorld-CR). We also introduce the \textit{Causal Cartographer Reasoning agent} (CTG-Reason) and evaluate it on the generated causal inference tasks.

\subsection{Counterfactual Matching}

We aim to infer the value of a target causal variable given observations or counterfactual evidence. To build such queries from CausalWorld, we use \textbf{counterfactual matching}. In this section, we define this concept and show how it allows building real world counterfactuals. 
First, we must establish a \textbf{causal blanket}, i.e. a set of variables that, if known, fully determine the target variable. The causal blanket differs from a Markov blanket \citep{pearl2014probabilistic} because it only includes direct paths to the target variable and does not necessarily require the nodes in the blanket to be the target's parents. For example, in the chain $A \rightarrow B \rightarrow C \rightarrow D$, $A$ forms a causal blanket for $C$. We assume that we can construct causal blankets from the CausalWorld graph. We formally define this concept in Definition~\ref{def:blanket}.

\begin{definition}[Causal Blanket]
    \label{def:blanket}
    Let $\mathcal{G}$ be a directed graphical model over a set of random variables $\mathcal{V}$, and let $T \in \mathcal{V}$ be a target variable. $\mathbf{anc}_{\mathcal{G}}(T)$ is the set of ancestors of $T$ with respect to $\mathcal{G}$.
    A set of variables $\mathcal{B} \subseteq \mathcal{V} \setminus \{T\}$ is called a \emph{causal blanket} for $T$ in $\mathcal{G}$ iff, $\mathcal{B} \subseteq \mathbf{anc}_{\mathcal{G}}(T)$ and conditioned on $\mathcal{B}$, the target variable $T$ is fully determined; that is, there exists a deterministic function $f$ such that
    \[
    T = f(\mathcal{B}).
    \]
    Equivalently, knowing $\mathcal{B}$ renders $T$ conditionally independent of all other variables in $\mathcal{V} \setminus (\mathcal{B} \cup \{T\})$. Unlike the Markov blanket, a causal blanket requires that $\mathcal{B}$ contains only variables with direct causal paths to $T$, but these need not be limited to the parents of $T$. It follows that a variable can have multiple causal blankets.
\end{definition}

It is not generally possible to obtain the true value of a counterfactual in the real world. Therefore, we focus on a subset of counterfactual queries accessible via \textbf{K-Matching}: finding two worlds within our observations that can act as factual and counterfactual worlds. I.e. observing the factual world and intervening on $K$ observed variables of the counterfactual world is equivalent to observing the counterfactual world. Figure \ref{fig:counterfactual-matching} illustrates the idea. We define this concept more formally below:

\begin{definition}[K-Matching]
Let $\mathcal{O}_o$ and $\mathcal{O}_c$ denote two sets of observations sampled from $\mathcal{V}$. Suppose that a target variable $T$ is present in both sets and that there exists a subset of shared observed observations $\mathcal{O}_s = \mathcal{O}_o \cap \mathcal{O}_c$ with $\mathcal{O}_s \neq \emptyset$. Let $\mathcal{B}_c \subseteq \mathcal{O}_c$ denote a causal blanket for $T$, with $|\mathcal{B}_c| = N$. We say that $\mathcal{B}_c$ can be \emph{$K$-matched} with $\mathcal{O}_o$ over $T$ if it is possible to build a new causal blanket for $T$ with $K$ interventions from $\mathcal{B}_c \setminus \mathcal{O}_s$ and $N - K$ observations from $\mathcal{O}_s$.
\end{definition}

\begin{theorem}[K-Matching Equivalence]
\label{thm:k-matching}
Suppose that a causal blanket $\mathcal{B}_c$ is \emph{K-matched} with $\mathcal{O}_o$ over a variable $T$ and that $\mathcal{O}_o \setminus \mathcal{O}_s$ forms a causal blanket over each variable of $\mathcal{O}_s$. Then, observing $\mathcal{B}_c$ or observing $\mathcal{O}_o \setminus \mathcal{O}_s$ and intervening on $\mathcal{B}_c \setminus \mathcal{O}_s$ yields the same distribution for $T$; that is,
\[
P(T \mid do(\mathcal{B}_c \setminus \mathcal{O}_s), \mathcal{O}_o \setminus \mathcal{O}_s) = P(T \mid \mathcal{B}_c).
\]
\end{theorem}

The proof is given in Appendix \ref{app:proof_k_matching}. This theorem implies that, if we can find a causal blanket for the counterfactual world and determine the value of $N - K$ variables of the blanket from the factual world, then we can build a counterfactual with $K$ interventions over the remaining variables.

\subsection{Matching with the CausalWorld-CR Dataset}

The CausalWorld-CR dataset is divided into two query types: \textit{observations} and \textit{counterfactuals}. For both subsets, the inference task to solve consists of inferring the value of a target causal variable given causal ancestors. 
\textbf{Observation queries} are created by sampling target variables and their causal blanket from a single world. The reasoning model is evaluated by inferring the value of the target variable given the values of the blanket variables. Since the query is built from a single world, the ground-truth is provided by the source document from which the world is built. The model must answer a query of the type $P(\text{Target} \mid \mathcal{B}_o)$.
We use K-matching to build \textbf{counterfactual queries}. The model must answer a query of the type $P(T \mid do(\mathcal{B}_c\setminus\mathcal{O}_s),\mathcal{O}_o\setminus\mathcal{O}_s)$. 
We focus on $1$-matching in our experiments. 

By default, the strategy defined above generates highly unbalanced datasets. The nodes that have a high degree and are present in multiple worlds are over-represented. We balance the dataset to reduce these effects and generate a dataset of 400 samples (see Appendix \ref{app:dataset_details} for details).

\subsection{CTG-Reason}

We build upon the methodology proposed by \citet{DBLP:journals/corr/abs-2410-06392} to create a graph-enhanced causal inference agent. The agent follows the steps of causal inference described in Section \ref{sec:background}. It performs step-by-step causal reasoning by computing the value of a variable from its direct causal parents only, or when anticausal reasoning is needed, from its direct children. At each step, the LLM's context only has access to the required parent/children variables.
This approach, illustrated in Figure \ref{fig:step-by-step-cr}, respects causal constraints and ensures that the agent only uses causal information for its reasoning, increasing efficiency and robustness by alleviating dependencies on non-causal and spurious correlations. An complete overview with the extraction agent is also shown in Figure \ref{fig:full-pipeline} of Appendix \ref{app:full-overview}.
The agent executes the plan described on the next page. 

\begin{enumerate}
    \item \textbf{Abduction:} If a twin graph is provided (with factual and counterfactual worlds), the agent performs abduction in an anticausal manner and computes the value of the exogenous variables of the factual world from their children. If their children are not observed, they are inferred from their own children \textit{recursively}. The values for the exogenous variables are transferred to the counterfactual world.
    \item \textbf{Intervention:} The counterfactual world is intervened upon: the incoming edges to all intervened variables are removed and the node values are fixed by the intervention value.
    \item \textbf{Prediction:} The target value is inferred from its parents. If they are not observed, they are inferred from their own parents \textit{recursively}. Since the set of observed and intervened variables forms a causal blanket, the target value is fully explained.
\end{enumerate}

\subsection{Evaluation}

We perform experiments with widely used LLMs tailored for reasoning (o3-mini \citep{openai2023o3mini}) and general purpose tasks (GPT-4.1 \citep{openai2025gpt41} and LLaMA-3.1-8B \citep{meta2024llama31}). We compare our step-by-step strategy against CausalCoT \citep{DBLP:conf/nips/JinCLGKLBAKSS23}, a chain-of-thought prompting strategy \citep{DBLP:conf/nips/Wei0SBIXCLZ22} for causal inference.
Since the evaluation queries are built from automatically scrapped real-world events, they express a diversity of types. The target variables to be predicted can be boolean, numerical or qualitative assessments of a trend. Moreover, the LLM may not provide a response corresponding to the same type. For instance, we find that LLMs tend to favor qualitative to numerical assessments (see Appendix \ref{app:additional_results} for more details).

\paragraph{Performance}

We look at the accuracy for boolean and trend queries as they correspond to categorical variables (true/false and increasing/decreasing/stable, respectively). Figure \ref{fig:trend-res} shows that both strategies yield similar results, although CTG-Reason achieves slightly better accuracy. We also note that LLaMA-3.1 was not able to complete the queries with the CausalCoT strategy. We discuss this aspect in the next paragraph.
Figure \ref{fig:numeric-err} shows the distribution of the relative error between the ground truth and the prediction for numerical counterfactual queries (in \% of the ground truth value), for o3-mini and GPT-4.1. We look at the relative error instead of the absolute difference to take into account the unit difference between queries. We also excluded outliers ($\sim$4\% of the answers were nonsensical numbers). We observe that CTG-Reason and CausalCoT strategies yield similar distributions for GPT-4.1 but that CTG-Reason achieves much lower error with o3-mini (curiously, for observation queries, CoT achieves better performance, see Figure \ref{fig:obs-numeric-err} of Appendix \ref{app:additional_results}).
Finally, we take a look at the semantic similarity between predicted and ground-truth answers in Figure \ref{fig:text-semantic-sim}. Since CTG-Reason does not have access to non-causal contextual information, its answer wording greatly differs from the ground-truth, although this is not indicative of an incorrect answer.

\begin{figure}
    \begin{tikzpicture}
    \pgfplotsset{
        mybarplot/.style={
            ybar,
            bar width=6pt,
            width=4.2cm,
            height=3.4cm,
            enlarge x limits=0.9,
            symbolic x coords={CTG,CoT},
            xtick=data,
            ylabel style={font=\footnotesize},
            xlabel style={font=\footnotesize},
            tick label style={font=\footnotesize},
            title style={font=\footnotesize,align=center},
            ymajorgrids = true,
            axis y line*=left,
            axis x line*=bottom,
            error bars/y dir=both, error bars/y explicit,
        }
    }
    \begin{groupplot}[
        group style={
            group size=4 by 2,
            horizontal sep=0.5cm,
            vertical sep=0.6cm,
        },
    ]
    
    \nextgroupplot[
        mybarplot,
        ymin=0,
        ylabel={Count},
        title={Steps},
    ]
    \addplot[fill=green!60,draw=black]  coordinates {(CTG,3.1) (CoT,1.925)};
    \addplot[fill=red!60,draw=black] coordinates {(CTG,3.6) (CoT,4.275)};
    \addplot[fill=cyan!60,draw=black] coordinates {(CTG,1.49) (CoT,3.5)};
    
    \nextgroupplot[
        mybarplot,
        ymin=0,
        title={Retries},
    ]
    \addplot[fill=green!60,draw=black]  coordinates {(CTG,0.1) (CoT,0.925)};
    \addplot[fill=red!60,draw=black] coordinates {(CTG,0.6) (CoT,3.275)};
    \addplot[fill=cyan!60,draw=black] coordinates {(CTG,0.2425) (CoT,2.5)};
    
    \nextgroupplot[
        mybarplot,
        ymin=0,
        title={Input Tokens},
    ]
    \addplot[fill=green!60,draw=black]  coordinates {(CTG,4390) (CoT,9695) };
    \addplot[fill=red!60,draw=black] coordinates {(CTG,6480) (CoT,16766)};
    \addplot[fill=cyan!60,draw=black] coordinates {(CTG,7801) (CoT,27588)};
    
    \nextgroupplot[
        mybarplot,
        ymin=0,,ymax=10001,
        title={Output Tokens},
    ]
    \addplot[fill=green!60,draw=black]  coordinates {(CTG,1188) (CoT,5654)};
    \addplot[fill=red!60,draw=black] coordinates {(CTG,700) (CoT,1915) };
    \addplot[fill=cyan!60,draw=black] coordinates {(CTG,807) (CoT,8879)};
    
    \end{groupplot}

    \tikzset{
        shadowed/.style={preaction={transform canvas={shift={(1pt,-1pt)}},draw=black!60,very thick}},
      }
    \draw[ultra thick, shadowed,-latex,ForestGreen] (8.29,1.7) to node[above left] {\textbf{-72\%}} (7.35,0.46);
    \draw[ultra thick, shadowed,-latex,ForestGreen] (11.425,1.65) to node[above left] {\textbf{-91\%}} (10.455,0.155);
    
    \path (group c1r1.south west)--(group c4r1.south east) coordinate[pos=0.5] (legendpos);
    \node[below=0.5cm of legendpos,font=\footnotesize] {
        \begin{tabular}{cccc}
        \raisebox{2.5pt}{\tikz{\draw[fill=green!60,draw=black] (0,0) rectangle (7pt,7pt);}} o3-mini
        \hspace{10pt}
        \raisebox{2.5pt}{\tikz{\draw[fill=red!60,draw=black] (0,0) rectangle (7pt,7pt);}} GPT-4.1
        \hspace{10pt}
        \raisebox{2.5pt}{\tikz{\draw[fill=cyan!60,draw=black] (0,0) rectangle (7pt,7pt);}} LLaMA-3.1-8B
        \end{tabular}
    };
    \end{tikzpicture}
    \caption{Statistics on the model answers with CTG-Reason and CausalCoT. (left) The average number of steps (i.e. model calls) required to solve a query. (middle left) The average number of retries after a failure to give a response (i.e. incorrect code formatting). (middle right) The average number of input tokens for the model. (right) The average number of output tokens.   }
    \label{fig:efficiency-results}
\end{figure}
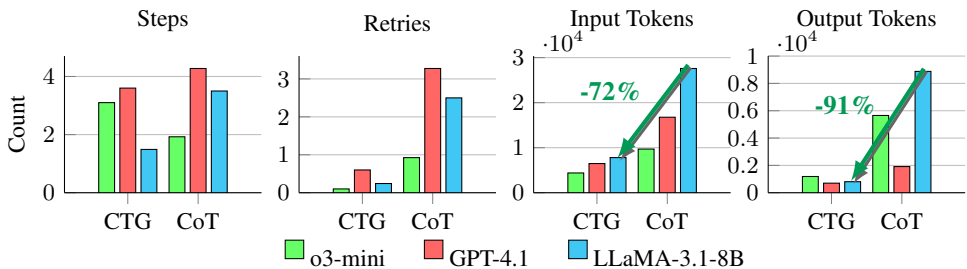

\paragraph{Efficiency}
We study the efficiency of our proposed model in Figure \ref{fig:efficiency-results}. While CTG-Reason decomposes the problem into multiple steps and requires several LLM calls to solve a problem, this is balanced by a significantly lower quantity of retries. It can be explained by the restriction of the context to causal components, which reduces the scope of the problem and the size of the context window, making the problem \textit{tractable}. As seen in the previous paragraph, this does not come at the cost of a reduced performance and can greatly help small models, as illustrated with LLaMA-3.1-8B: its context size is reduced by 72\% and the output size is reduced by 91\%.

\begin{figure}
    \begin{subfigure}[b]{0.32\linewidth}
        \begin{tikzpicture}
            \begin{axis}[
                ybar,
                bar width=6pt,
                ymin=0,
                ymax=1.1,
                ylabel={Accuracy (bool)},
                symbolic x coords={o3-mini,GPT-4.1,~~~LLaMA-3.1},
                xtick=data,
                width=\linewidth,
                height=3cm,
                legend style={at={(1.0,0.96)},anchor=south east,legend columns=-1},
                enlarge x limits=0.25,
                every axis/.append style={font=\tiny}
            ]
            \addplot+[fill=blue!60,draw=black] coordinates {(o3-mini,1) (GPT-4.1,1) (~~~LLaMA-3.1,0.81)};
            \addplot+[fill=orange!75!yellow!80,draw=black]  coordinates {(o3-mini,0.83)(GPT-4.1,0.96) (~~~LLaMA-3.1,0)};
            \end{axis}
            \node (s) at (2.55, 0.1) {*};
        \end{tikzpicture}
        \begin{tikzpicture}
            \begin{axis}[
                ybar,
                bar width=6pt,
                ymin=0,
                ymax=1.1,
                ylabel={Accuracy (trend)},
                symbolic x coords={o3-mini,GPT-4.1,~~~LLaMA-3.1},
                xtick=data,
                width=\linewidth,
                height=3cm,
                legend style={at={(1.0,1)},anchor=north east,legend columns=-1},
                enlarge x limits=0.25,
                every axis/.append style={font=\tiny}
            ]
            \addplot+[fill=blue!60,draw=black] coordinates {(o3-mini,0.61) (GPT-4.1,0.56) (~~~LLaMA-3.1,0.52)};
            \addplot+[fill=orange!75!yellow!80,draw=black] coordinates {(o3-mini,0.58) (GPT-4.1,0.55) (~~~LLaMA-3.1,0)};
            \legend{CTG-R.,C.CoT}
            \end{axis}
            \node (s) at (2.55, 0.1) {*};
        \end{tikzpicture}
        \caption{Accuracy for bool/trend queries. }
        \label{fig:trend-res}
    \end{subfigure}
    \hfill
    \begin{subfigure}[b]{0.32\linewidth}
        \includegraphics[width=\linewidth]{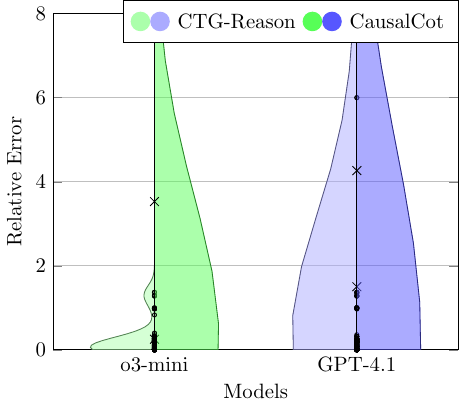}
        \caption{Relative error for num. queries. }
        \label{fig:numeric-err}
    \end{subfigure}
    \hfill
    \begin{subfigure}[b]{0.34\linewidth}
        \includegraphics[width=\linewidth]{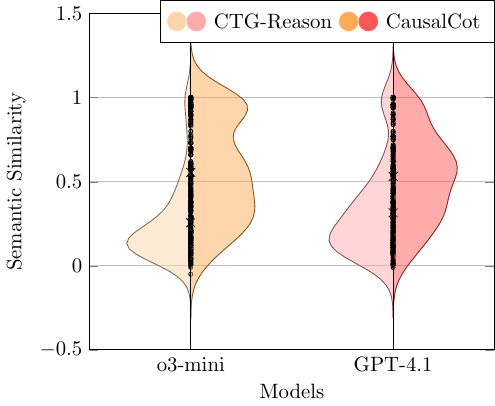}
        \caption{Cosine similarity for text queries. }
        \label{fig:text-semantic-sim}
    \end{subfigure}
    \caption{(a) Results on the boolean and trend subsets of CausalWorld-CR. Results are shown for o3-mini, GPT-4o and LLaMA-3.1-8B, using CTG-Reason (left bar/half) and CausalCoT (right bar/half). (b) and (c) Violin plots of the relative error (in \%) and semantic (cosine) similarity between numerical/text ground truth and predicted answers on the counterfactual set. (*) The majority of queries with LLaMA-3.1-8B-CausalCoT returned with a timeout. }
    \label{fig:main-results}
\end{figure}
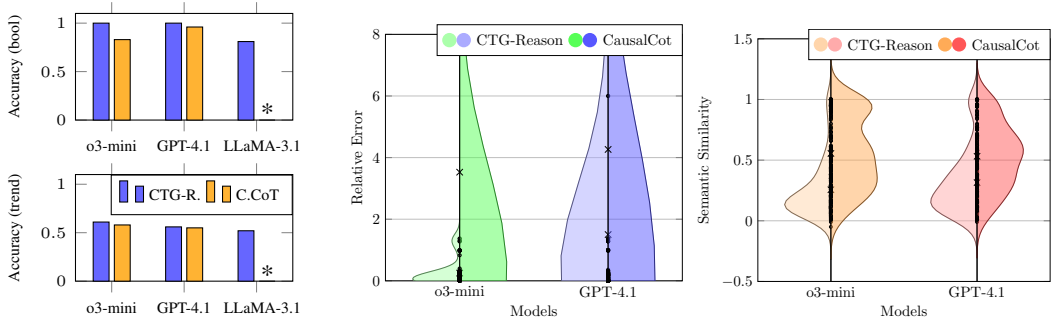

\section{Related Work}

Traditional causal extraction methods rely on knowledge-based or statistical methods which can be brittle when causal relationships spread across multiple sentences or paragraphs \citep{DBLP:journals/kais/YangHP22}.  
More recent approaches rely on deep learning architectures and transformers, notably CausalBERT \citep{DBLP:conf/sai/KhetanRASF21} which relies on BERT \citep{DBLP:conf/naacl/DevlinCLT19}, but the use of larger LLMs remains under-explored.  
CauseNet \citep{DBLP:conf/cikm/HeindorfSWNP20} is a database of causal relationships extracted from the web by mining linguistic patterns, but it does not make the distinction between variable and value that allows us to do counterfactual matching.
Our NxGraphRAG method also differs from existing graph-RAG systems (e.g. GraphRAG \citep{DBLP:journals/corr/abs-2404-16130}) as we do not use knowledge graphs but causal graphs. Similarly to \citet{DBLP:conf/emnlp/0008YTBYLCHDDLS24} and \citet{DBLP:journals/corr/abs-2502-13025}, we integrate the graph-RAG approach to iteratively build a graph \citep{DBLP:journals/corr/abs-2408-08921}.
Studies have evaluated the causal reasoning abilities of LLMs, notably \citep{DBLP:journals/tmlr/ZecevicWDK23}, \citep{DBLP:conf/iclr/Jin0LPSMDS24} and Cladder \citep{DBLP:conf/nips/JinCLGKLBAKSS23}: a comprehensive benchmark for evaluating causal reasoning in language. Unlike in our work however, Cladder is built synthetically.

\section{Limitations}
\label{sec:limitations}

The iterative nature of the causal extraction implies that the processing order of the documents has an impact on the final causal network. This is not a desired behavior as the causal relationships should be order-agnostic, potentially leading to a suboptimal configuration of the network. 
Moreover, since our counterfactual reasoning framework relies on causal blankets, we must assume knowledge of the full causal graph. It is not possible to guarantee that this is the case in the real world.
Numerical estimations are also challenging for LLMs, which often output biased results \citep{mirzadeh2024gsm}, limiting the interpretation of a direct comparison.
While our method aims to reduce bias and improve robustness by retaining causal relationships instead of (potentially spurious) correlations, we rely on the accuracy and honesty of the source data and are sensitive to adversarial attacks and misinformation injections.
We discuss further the broader impact and ethical considerations of our work in Appendix~\ref{app:impact}.

\section{Conclusion}
\label{sec:conclusion}

We introduce the \textbf{Causal Cartographer}, a twofold framework composed of causal extraction and inference agents that learn causal knowledge from natural language. We use this framework to build a network of causal knowledge and prove that it allows the estimation of real-world counterfactuals. 
We show that our proposed step-by-step inference agent can outperform chain-of-thought baselines on counterfactual reasoning and greatly reduce the inference cost by alleviating the impact of non-causal information.
We hope that our work will inspire the creation of more robust and efficient reasoning agents based on causal principles. In particular, allowing agents to learn from counterfactual information is a promising direction towards building systems of higher cognition.


\bibliography{ref}
\bibliographystyle{iclr2025_conference.bst}

\newpage
\appendix

\section{Broader Impact and Ethical Considerations}
\label{app:impact}

This work presents a method for building a repository of causal knowledge from real-world data in natural language, more specifically from news articles, and estimating counterfactual outcomes. We expect that it can help build better reasoning systems able to perform causal inference in natural language. The grounding in real-world data can particularly help improve the abilities of large language models on many downstream real-world tasks. Causal inference has been argued as a promising direction to reduce bias and increase fairness, trustworthiness and safety of AI systems \citep{pearl2009causality,DBLP:journals/corr/abs-2011-15091,DBLP:journals/pieee/ScholkopfLBKKGB21,DBLP:books/acm/22/BareinboimCII22,DBLP:conf/cvpr/MaoXW0YBV22,DBLP:journals/corr/abs-2302-00293,DBLP:conf/ijcai/Gendron0WD24,DBLP:conf/emnlp/GendronNPWD24,DBLP:conf/iclr/RichensE24} and we hope that this work can foster research in this direction and help bridge the gap with real-world applications.

However, we acknowledge that our work can also have negative impact. First, as mentioned in the limitations (Section \ref{sec:limitations}), we rely on the accuracy and honesty of the source data and are not robust to adversarial attacks. Our method gives equal importance to all causal relationships, which means that a small amount of adversarial data can have a significant impact on the extraction and inference, exacerbating harmful trends and heavily affecting the downstream predictions. As a consequence, we do not recommend using our method on potentially unreliable sources. Second, our work can be used to make counterfactual predictions about real-world situations. While this work is only a first step in this direction, we expect that it will lead to economical and societal prediction engines. For instance, companies could make predictions about the economic impact of a businesses decision, or government organizations could predict the future impact of a policy on a population. Such usage can have very high impact over many areas and populations. It could provide guidance for enforcing better policies improving the well-being of populations but can also be used by malicious actors for harm. Even without the intervention of such actors, an over reliance on such prediction tools without critical analysis could similarly lead to disastrous effects in case of errors or misinterpretation of the prediction.

In its current stage, our work does not presents any such risks of misuse. However, we would like to emphasize the ethical considerations and risks links to the pursuit of this research direction and argue that future work should keep them into consideration.

\section{Overview of the Entire Causal Cartographer Pipeline}
\label{app:full-overview}

\begin{figure}
    \centering
    \includegraphics[width=\linewidth]{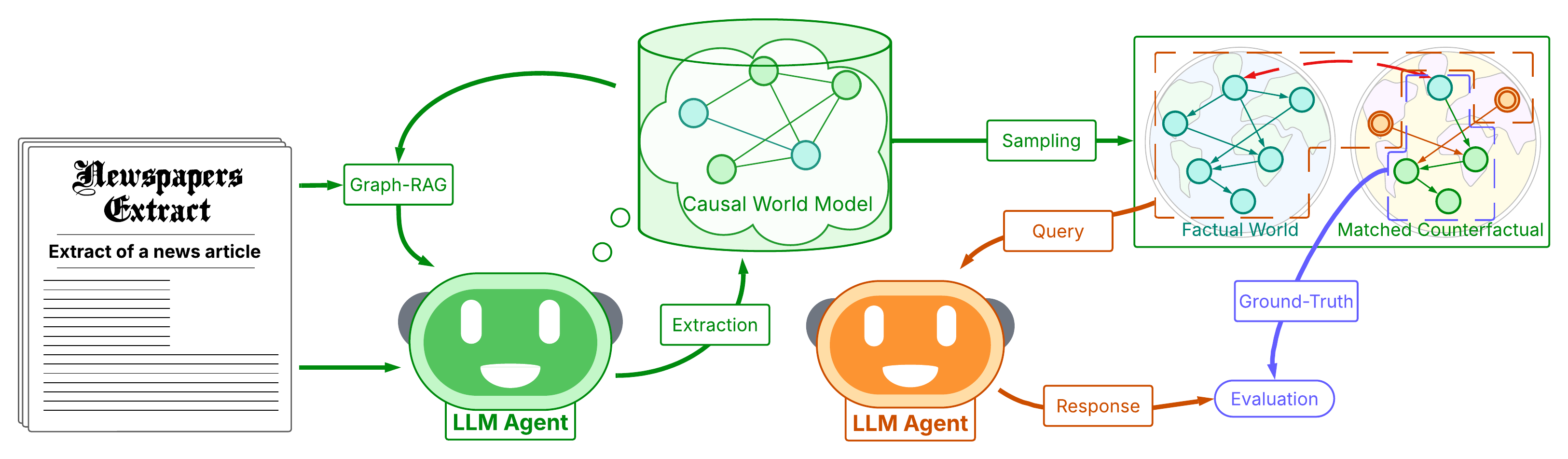}
    \caption{Overview of the complete Causal Cartographer pipeline with extraction and reasoning agents. The title, content and metadata from a news source is provided to the context of the extraction agent. It then generates causal variables and their relationships and adds them to the CausalWorld network, making explicit the causal knowledge of the framework. A graphRAG system further compares article data and agent node suggestions and adds them to the extraction agent's context to guide the extraction process. Then, ground-truth counterfactuals are sampled from the CausalWorld graph using matching: observations in the counterfactual world that match those in a factual world are removed and replaced with the factual world for abduction. The reasoning agent is evaluated on the generated real-world counterfactual queries. }
    \label{fig:full-pipeline}
\end{figure}

Figure \ref{fig:full-pipeline} shows an unified representation of the Causal Cartographer with both extraction and reasoning agents. The CausalWorld network is the intermediate representation connecting the two aspects of the framework: representing the extracted knowledge and sampling counterfactual questions.

\section{Proof of Theorem \ref{thm:k-matching}}
\label{app:proof_k_matching}

In this section, we prove the \textbf{K-Matching} theorem (Theorem \ref{thm:k-matching}) stated in Section \ref{sec:counterfactual-inference}.

\begin{proof}

We suppose that a causal blanket $\mathcal{B}_c$ is \emph{K-matched} with $\mathcal{O}_o$ over a variable $T$ and that $\mathcal{O}_o \setminus \mathcal{O}_s$ forms a causal blanket over each variable of $\mathcal{O}_s$. We aim to show that $P(T \mid do(\mathcal{B}_c \setminus \mathcal{O}_s), \mathcal{O}_o \setminus \mathcal{O}_s) = P(T \mid \mathcal{B}_c)$, i.e. that a counterfactual query following this set of assumptions can be rewritten as an observational query.

We first rewrite the counterfactual query as a probability distribution over the set of exogenous variables $\mathbf{U}$, following Equation \ref{eq:counterfactual-exogenous} \citep{pearl2009causality}:

\begin{align*}
    P(T \mid do(\mathcal{B}_c \setminus \mathcal{O}_s), \mathcal{O}_o \setminus \mathcal{O}_s) &= \sum_{U \in \mathbf{U}} P(T \mid do(\mathcal{B}_c \setminus \mathcal{O}_s),U)P(U \mid \mathcal{O}_o \setminus \mathcal{O}_s)  
\end{align*}

$\mathcal{B}_c = (\mathcal{B}_c \setminus \mathcal{O}_s) \cup \mathcal{B}_c$ forms a causal blanket over $T$. Therefore, $T \independent \mathbf{U} \mid \mathcal{B}_c$. The only paths from $\mathbf{U}$ to $T$ must go through the variables instantiating $\mathcal{O}_s$. We call them $\mathcal{S}$. We infer that $T \independent \mathbf{U} \setminus \mathcal{S} \mid \mathcal{B}_c \setminus \mathcal{S}$. We can then rewrite the equation as follows:

\begin{align*}
    P(T \mid do(\mathcal{B}_c \setminus \mathcal{O}_s), \mathcal{O}_o \setminus \mathcal{O}_s) &= \sum_{\mathcal{S} \in \mathbf{U}} P(T \mid do(\mathcal{B}_c \setminus \mathcal{O}_s),\mathcal{S})P(\mathcal{S} \mid \mathcal{O}_o \setminus \mathcal{O}_s)  
\end{align*}

We now show that $\mathcal{O}_s$ is the only possible instantiation of $\mathcal{S}$. It is assumed that $\mathcal{O}_o \setminus \mathcal{O}_s$ forms a causal blanket over $\mathcal{S}$. Therefore, $\mathcal{O}_s$ can be deterministically computed from $\mathcal{O}_o$ and the probability $P(\mathcal{S} \mid \mathcal{O}_o \setminus \mathcal{O}_s)$ will return zero probability except for the values $\mathcal{O}_s$. the equation can then be rewritten as follows:

\begin{align*}
    P(T \mid do(\mathcal{B}_c \setminus \mathcal{O}_s), \mathcal{O}_o \setminus \mathcal{O}_s) &= P(T \mid do(\mathcal{B}_c \setminus \mathcal{O}_s),  \mathcal{O}_s)  
\end{align*}

Rule 2 of \textit{do-calculus} \citep{pearl2009causality, DBLP:conf/uai/Pearl12} states that an intervention can be reduced to an observation if no backdoor path connects the intervened variable to the variable of interest. It can be written as follows: 

\begin{equation}
    P(y|do(x),do(z), w) = P(y|do(x), z, w) \text{~if~} (Y \independent Z|X, W)_{\mathcal{G}_{\overline{X}\underline{Z}}}
    \label{eq:do_rule2}
\end{equation}

Since $do(\mathcal{B}_c \setminus \mathcal{O}_s),  \mathcal{O}_s$ forms a causal blanket over $T$, rule 2 applies and the equation can be further simplified into the desired quantity:

\begin{align*}
    P(T \mid do(\mathcal{B}_c \setminus \mathcal{O}_s), \mathcal{O}_o \setminus \mathcal{O}_s) &= P(T \mid do(\mathcal{B}_c \setminus \mathcal{O}_s),  \mathcal{O}_s) \\
    &= P(T \mid (\mathcal{B}_c \setminus \mathcal{O}_s),  \mathcal{O}_s) \\
    &= P(T \mid \mathcal{B}_c)
\end{align*}

\end{proof}

\section{Details on the NxGraphRAG Architecture}
\label{app:rag_details}

\begin{figure}
    \centering
    \includegraphics[width=0.9\linewidth]{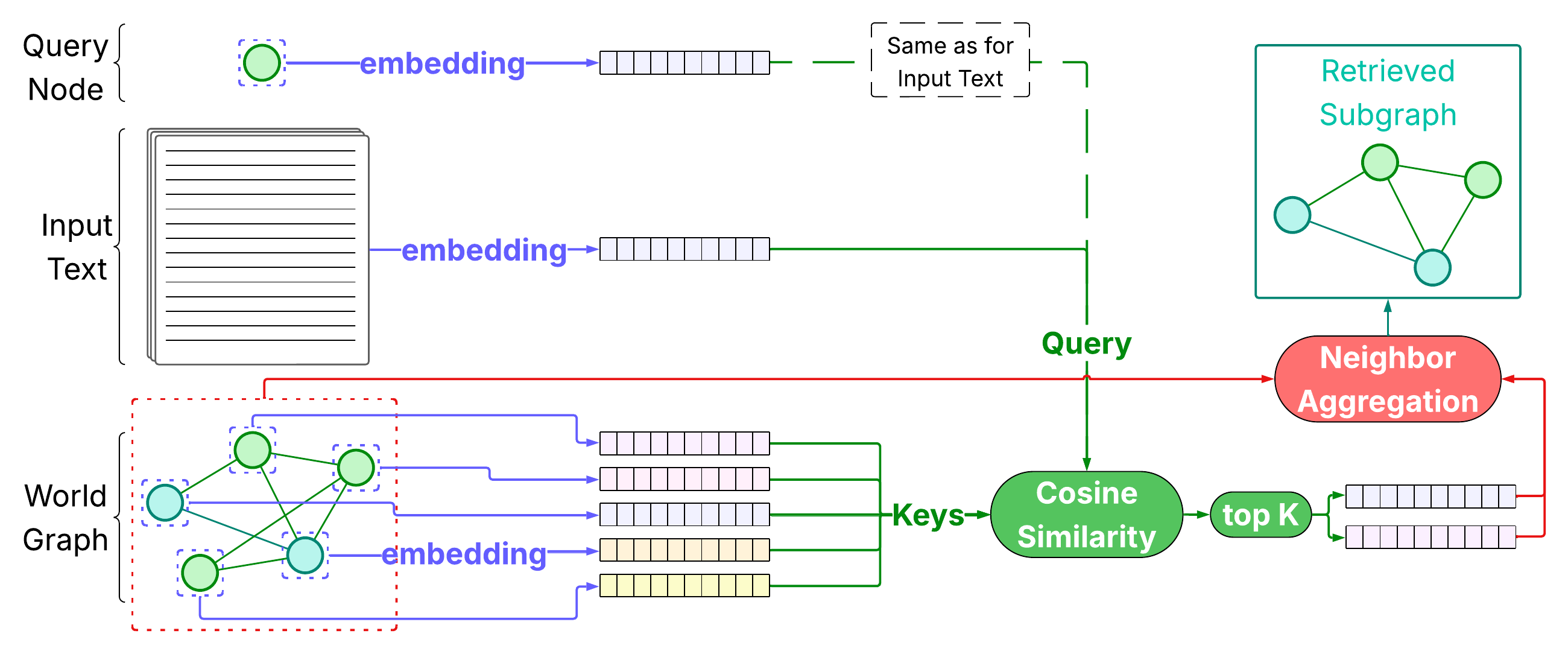}
    \caption{Overview of the NxGraphRAG pipeline. The attributes of the graph database are embedded and stored in a vector database. Depending on the use case, an input document or a candidate node are embedded as a query vector. The query vector is compared against the graph node (key) vectors using cosine similarity. The $K$ closest embeddings are selected. For each node, its $P$-level neighbors are also selected. The subgraph is returned and added to the agent context.   }
    \label{fig:NxGraph-RAG}
\end{figure}

This section provides additional details regarding the graph retrieval-augmented generation pipeline introduced in Section \ref{sec:causal_extraction}. We introduce Networkx Graph Retrieval-Augmented Generation (NxGraphRAG), jointly used with the proposed Causal Cartographer Extraction agent (CTG-Extract). An overview of the method is shown in Figure \ref{fig:NxGraph-RAG}. We use Langchain \footnote{https://www.langchain.com/} to build the RAG pipeline. It is provided to the agent as a callable tool.

NxGraphRAG generates embeddings for the world graph nodes, candidate nodes proposed by the agent, and input documents, that can be compared using cosine similarity. We use OpenAI Embeddings \citep{DBLP:journals/corr/abs-2201-10005} to generate the embeddings. For the nodes, all the attributes are provided as a list of key-value character chains. The embeddings of the world graph represents the keys against which the queries, i.e. the candidate nodes and input documents, are compared with. While query vectors are generated on the fly, key vectors are stored in an vector database. We use an in-memory database because the generated world graph is small enough to fit in the memory (see Section \ref{sec:causal_world} for more details). NxGraphRAG returns the top-$K$ most similar nodes with the query. It also returns the neighbors of these nodes, up to a level $P$ neighborhood. We use $K=3$ and $P=2$ to balance exhaustiveness and efficiency. 

\section{Details on the Structure of CausalWorld}
\label{app:causalworld_structure}

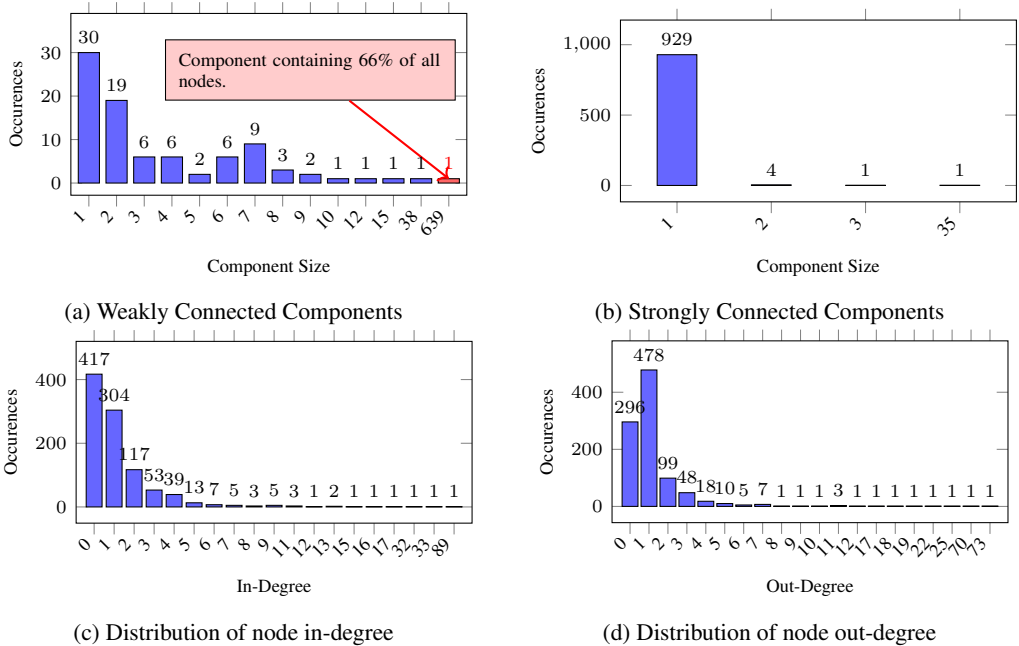
\begin{figure}[htbp]
    \centering
    \begin{subfigure}[b]{0.49\textwidth}
        \centering
        \begin{tikzpicture}
        \begin{axis}[
        ybar,
        ymax=39,
        bar width=8pt,
        width=\linewidth,
        height=4cm,
        xlabel={Component Size},
        ylabel={Occurences},
        symbolic x coords={1,2,3,4,5,6,7,8,9,10,12,15,38,639},
        xtick={1,2,3,4,5,6,7,8,9,10,12,15,38,639},
        xticklabel style={rotate=45, anchor=east},
        nodes near coords,
        enlarge x limits=0.05,
        /pgf/bar shift=0pt,
        every axis/.append style={font=\scriptsize}
        ]
        \addplot[fill=blue!60,draw=black] coordinates {
        (1,30) (2,19) (3,6) (4,6) (5,2) (6,6)
        (7,9) (8,3) (9,2) (10,1) (12,1) (15,1)
        (38,1)
        };
        \addplot[red,fill=red!60,draw=black] coordinates {
            (639,1)
            };
            
        \coordinate (redbar) at (axis cs:639,1);
        \node[draw=black, fill=red!20, inner sep=5pt, text width=3.5cm] (dialog)
          at (axis cs:9, 26) {Component containing 66\% of all nodes.};
        \draw[->, thick, red] (dialog) -- (redbar);
        \end{axis}
        \end{tikzpicture}
        \caption{Weakly Connected Components}
    \end{subfigure}
    \hfill
    \begin{subfigure}[b]{0.49\textwidth}
        \centering
        \begin{tikzpicture}
        \begin{axis}[
        ybar,
        ymax=1175,
        bar width=15pt,
        width=\linewidth,
        height=4cm,
        xlabel={Component Size},
        ylabel={Occurences},
        symbolic x coords={1,2,3,35},
        xtick=data,
        xticklabel style={rotate=45, anchor=east},
        nodes near coords,
        enlarge x limits=0.2,
        every axis/.append style={font=\scriptsize}
        ]
        \addplot[fill=blue!60,draw=black] coordinates {
        (1,929) (2,4) (3,1) (35,1)
        };
        \end{axis}
        \end{tikzpicture}
        \caption{Strongly Connected Components}
    \end{subfigure}
    
    \begin{subfigure}[b]{0.49\textwidth}
        \centering
        \begin{tikzpicture}
        \begin{axis}[
        ybar,
        ymax=520,
        bar width=6pt,
        width=\linewidth,
        height=4cm,
        xlabel={In-Degree},
        ylabel={Occurences},
        symbolic x coords={0,1,2,3,4,5,6,7,8,9,11,12,13,15,16,17,32,33,89},
        xtick=data,
        xticklabel style={rotate=45, anchor=east},
        nodes near coords,
        enlarge x limits=0.05,
        every axis/.append style={font=\scriptsize}
        ]
        \addplot[fill=blue!60,draw=black] coordinates {
        (16,1) (89,1) (0,417) (5,13) (9,5) (1,304) (4,39) (6,7) (33,1) (13,2) (3,53) (11,3) (12,1) (15,1) (17,1) (2,117) (8,3) (32,1) (7,5)
        };
        \end{axis}
        \end{tikzpicture}
        \caption{Distribution of node in-degree}
    \end{subfigure}
    \hfill
    \begin{subfigure}[b]{0.49\textwidth}
        \centering
        \begin{tikzpicture}
        \begin{axis}[
        ybar,
        ymax=580,
        bar width=6pt,
        width=\linewidth,
        height=4cm,
        xlabel={Out-Degree},
        ylabel={Occurences},
        symbolic x coords={0,1,2,3,4,5,6,7,8,9,10,11,12,17,18,19,22,25,70,73},
        xtick=data,
        xticklabel style={rotate=45, anchor=east},
        nodes near coords,
        enlarge x limits=0.05,
        every axis/.append style={font=\scriptsize}
        ]
        \addplot[fill=blue!60,draw=black] coordinates {
        (7, 7) (70, 1) (1, 478) (3, 48) (18, 1) (11, 3) (4, 18) (12, 1) (0, 296) (22, 1) (6, 5) (2, 99) (5, 10) (9, 1) (19, 1) (17, 1) (10, 1) (73, 1) (25, 1) (8, 1)
        };
        \end{axis}
        \end{tikzpicture}
        \caption{Distribution of node out-degree}
    \end{subfigure}
    
    \caption{Distribution of weakly and strongly connected components and node degrees in CausalWorld. The majority of nodes (66\%) belong to one connected component (highlighted in red) while 3\% are isolated and the others are in components of intermediate size. 5\% of vertices are in non-trivial strongly connected components and are part on feedback loops. The majority of nodes have less than five neighbors, further highlighting the sparsity of the graph.  }
    \label{fig:causal-world-components}
\end{figure}

Figure \ref{fig:causal-world-components} shows the distribution of strongly and weakly connected components in the CausalWorld graph, highlighting that the majority of the network is connected and forms a single component. Despite this structure, few feedback loops are observed.

\begin{figure}
        \centering
        \begin{tikzpicture}
        \begin{axis}[
        ybar,
        ymax=21,
        bar width=6pt,
        width=0.5\linewidth,
        height=4cm,
        xlabel={Cycle Length},
        ylabel={Occurences},
        symbolic x coords={2,3,4,5,6,7,8,9,10,11,12,13,14},
        xtick=data,
        xticklabel style={rotate=45, anchor=east},
        nodes near coords,
        enlarge x limits=0.05,
        every axis/.append style={font=\scriptsize}
        ]
        \addplot[fill=blue!60,draw=black] coordinates {
        (2, 10) (3, 14) (4, 17) (5, 15) (6, 16) (7, 14) (8, 15) (9, 17) (10, 13) (11, 10) (12, 5) (13, 2) (14, 1)
        };
        \end{axis}
        \end{tikzpicture}
        \caption{Distribution of cycles in CausalWorld. There are 149 cycles in the graph, with lengths comprised between two and 14. }
    \end{figure}
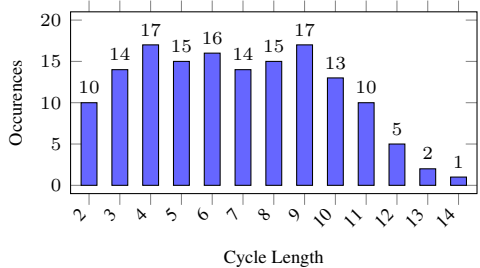

We now investigate the topics extracted from the news sources and represented in CausalWorld. We create clusters based on semantic embeddings. We only use the node attributes invariant to the current world. Due to the large scale of the graph, we use a smaller embedding model than in the NxGraphRAG pipeline. Since semantic clustering is only used for visualizing the graph content and is not connected to a downstream task, a downgraded performance would have no effect on the rest of our pipeline. We use all-mpnet-base-v2 \citep{all-mpnet-base-v2} with the SentenceTransformers library \footnote{https://www.sbert.net/} to generate the embeddings and K-Means to attribute clusters. We use the Silhouette score \citep{rousseeuw1987silhouettes} to determine the optimal number of clusters and find that the network can be best divided into 44 clusters. They can be observed in Figure \ref{fig:causal-world-graph-structure} of the main paper.

\begin{figure}
    \centering
    \includegraphics[width=\linewidth,trim=0cm 1cm 0cm 1cm,clip]{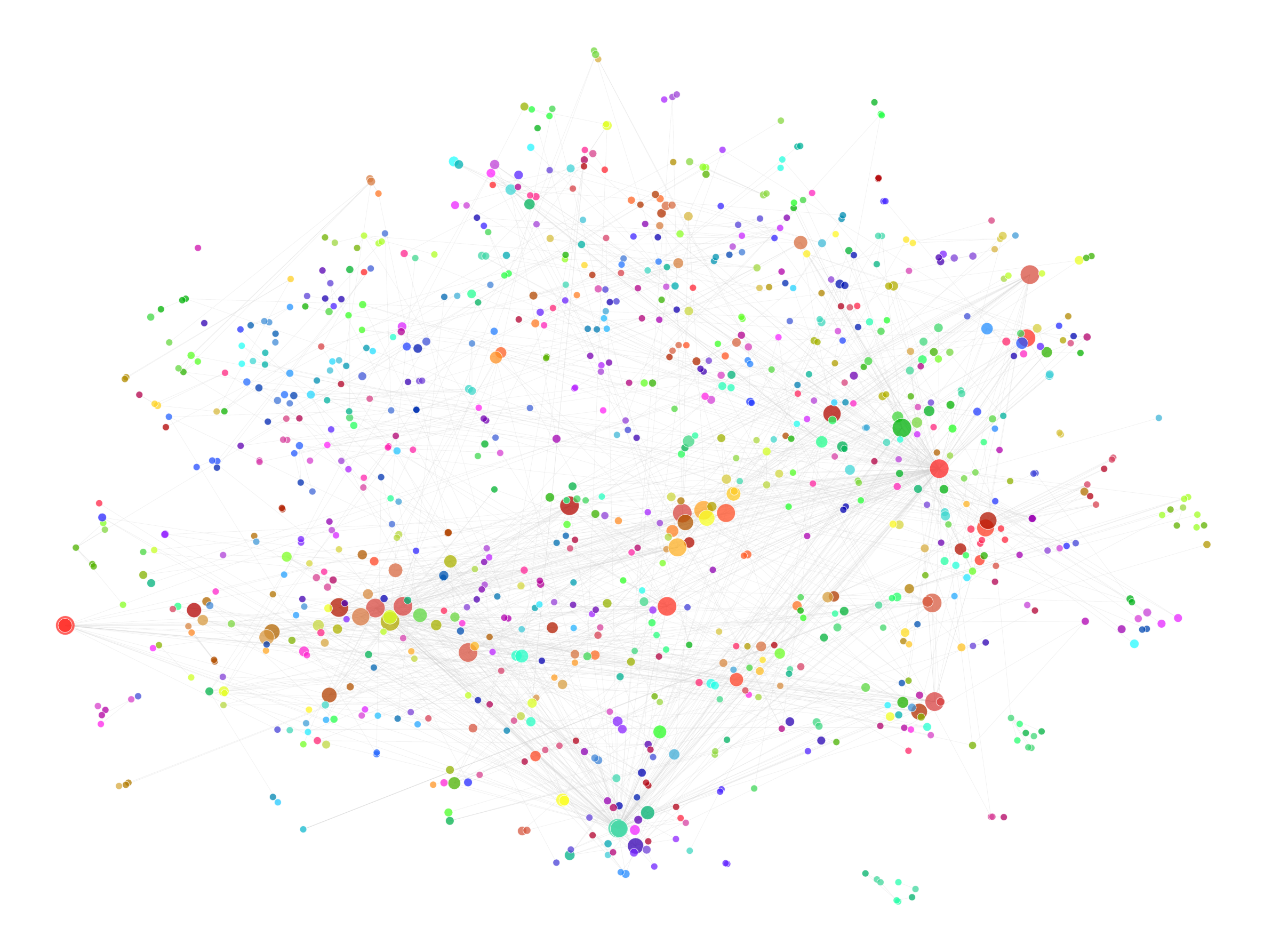}
    \caption{CausalWorld graph structure divided by Louvain communities. Colors correspond to the Louvain community in which belongs the node. Node size is based on the amount of worlds in which the node appear: more often a node appears, the bigger it is on the visualization. A total of 109 communities exists in the graph, therefore colors can correspond to several communities. This figure is for illustration purpose only. }
    \label{fig:causal-world-louvain}
\end{figure}

We further use the Louvain method \citep{blondel2008fast} to detect communities within the graph based on structural information only. We discover 109 communities within the network. Eight nodes are bridge nodes separating four communities or more: \texttt{Crude Oil Prices}, \texttt{Crude Oil Prices}, \texttt{Novel Coronavirus Outbreak}, \texttt{Global Economic Growth}, \texttt{US-China Trade Talks}, \texttt{Middle East Unrest}, \texttt{US Strike}, \texttt{US-Iran Flare-up} and \texttt{US-Iran Tensions Easing}. These nodes are key elements of the graph that enable the propagation of information from one community to the next. Figure \ref{fig:causal-world-louvain} illustrates the division of the network into communities.

\begin{figure}[htbp]
    \centering
    \begin{subfigure}[b]{\textwidth}
        \centering
        \begin{tikzpicture}
            \begin{axis}[
                xbar,
                bar width=15pt,
                enlarge y limits=0.07,
                xlabel={World Count},
                symbolic y coords={Crude Oil Prices,Novel Coronavirus Outbreak,Global Stock Performance,US Stock Market Performance,US Strike,Indian Rupee Exchange Rate,Dollar Index,Global Economic Growth,Safe Haven Assets,Gold Prices},
                ytick=data,
                height=5cm,
                nodes near coords,
                nodes near coords align={horizontal},
                xmin=0,
                xmax=280,
                grid=major,
                bar width=5pt,
                width=0.85\linewidth,
                every axis/.append style={font=\scriptsize},
                grid style={dashed}
                ]
                \addplot[fill=blue!60] coordinates {
                (243,Crude Oil Prices) (94,Novel Coronavirus Outbreak) (53,Global Stock Performance) (48,US Stock Market Performance) (47,US Strike) (43,Indian Rupee Exchange Rate) (40,Dollar Index) (36,Global Economic Growth) (34,Safe Haven Assets) (33,Gold Prices)
                };
            \end{axis}
        \end{tikzpicture}
        \caption{Top ten nodes with the highest representation among saved worlds. }
    \end{subfigure}
    \hfill
    \begin{subfigure}[b]{\textwidth}
        \centering
        \begin{tikzpicture}
            \begin{axis}[
                ybar,
                bar width=15pt,
                enlarge x limits=0.025,
                xlabel={World Count},
                ylabel={Number of Nodes},
                symbolic x coords={1,2,3,4,5,6,7,8,9,10,11,12,13,14,15,16,18,19,22,23,25,28,29,30,31,32,33,34,36,40,43,47,48,53,94,243},
                xtick=data,
                nodes near coords,
                nodes near coords align={vertical},
                ymin=0,
                ymax=700,
                height=5cm,
                grid=major,
                bar width=6pt,
                width=\linewidth,
                every axis/.append style={font=\scriptsize},
                grid style={dashed}
                ]
                \addplot[fill=green!60] coordinates {
                (243, 1) (94, 1) (53, 1) (48, 1) (47, 1) (43, 1) (40, 1) (36, 1) (34, 1) (33, 1) (32, 1) (31, 2) (30, 1) (29, 1) (28, 1) (25, 2) (23, 3) (22, 3) (19, 1) (18, 4) (16, 2) (15, 1) (14, 4) (13, 4) (12, 3) (11, 1) (10, 3) (9, 5) (8, 5) (7, 16) (6, 13) (5, 17) (4, 36) (3, 82) (2, 143) (1, 611)
                };
            \node[draw, fill=red!20, inner sep=5pt, text width=3.5cm] (dialog)
              at (axis cs:29, 200) {Top ten nodes (see bar plot above).};
            \coordinate (topbars) at (axis cs:43,80);
            \end{axis}
            \draw[thick,dashed,draw=blue!60] (8.9,0.075) rectangle (12.27,0.36);
            \draw[->, thick, red] (dialog) -- (topbars);
        \end{tikzpicture}
        \caption{Distribution of the nodes based on their number of appearances in worlds. }
    \end{subfigure}
    
    \caption{Distribution of the nodes among the worlds extracted from the source documents. The higher the world count, the most document a node appears in. The top plot shows the amount of worlds for the top ten nodes and the botom plot shows the distribution for the entire graph. }
    \label{fig:node_world_distribution}
\end{figure}
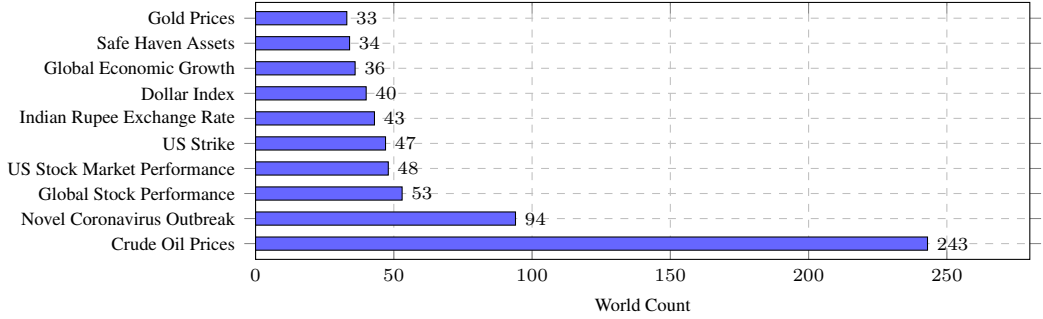
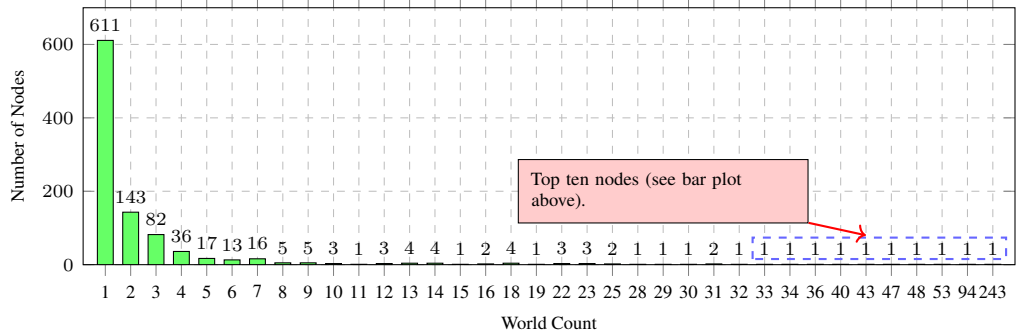

Figure \ref{fig:node_world_distribution} shows the distribution of nodes among the worlds extracted from the input documents. We can observe that the majority of the nodes belong to a single world but that 37\% of the nodes share two worlds or more, allowing information to be propagated across worlds. Since the majority of the nodes belong to the same component, most nodes can be used to compute counterfactuals.

\section{Additional Causal paths in CausalWorld}
\label{app:causal-paths}

\begin{figure}

    \begin{tikzpicture}[
        block/.style={
        rectangle, rounded corners,
        draw=black, very thick,
        font=\small,
        fill=blue!20,
        drop shadow, 
        minimum width=2cm, minimum height=1cm,
        align=center,
        inner sep=5pt,
        text width=2cm,
        },
        process/.style={
        rectangle, rounded corners,
        draw=black, very thick, dashed,
        font=\small,
        fill=red!20,
        drop shadow, 
        minimum width=2cm, minimum height=1cm,
        align=center,
        inner sep=5pt,
        text width=2cm,
        },
        causal arrow/.style={
        -{Stealth[length=4mm, width=3mm]},
        thick,
        shorten >=2pt
        },
        every node/.append style={on chain, node distance=1.5cm}
        ]
        \begin{scope}[start chain=going right]
            \node[block] (A) at (0,0) {Biofuel Demand};
            \node[process] (B) {Strong biofuel demand drives up palm oil prices.};
            \node[block] (C) {Palm Oil Prices};
            \node[process] (D) {Higher palm oil prices increase the vegetable oil price index.};
        \end{scope}
    
        \begin{scope}[start chain=going left, yshift=-1.75cm]
            \node[block] (E) at (0.815\linewidth,0) {FAO Vegetable Oil Price Index};
            \node[process] (F) {An increased vegetable oil price index raises overall food prices.};
            \node[block] (G) {Food Prices};
            \node[process] (H) {Higher food prices contribute to the overall rise in inflation.};
        \end{scope}
    
        \begin{scope}[start chain=going right, yshift=-3.5cm]
            \node[block] (I) at (0,0) {Inflation Rate};
            \node[process] (J) {Below-target inflation supports the decision to keep rates low.};
            \node[block] (K) {Fed Interest Rates};
            \node[process] (L) {Keeping rates low helps stimulate economic activity [...].};
        \end{scope}
    
        \begin{scope}[start chain=going left, yshift=-5.25cm]
            \node[block] (M) at (0.815\linewidth,0) {US Economic Growth};
            \node[process] (N) {US economic growth bolsters investor confidence [...].};
            \node[block] (O) {Global Stock Performance};
            \node[process] (P) {Global market performance influences China's stock market [...].};
        \end{scope}
    
        \begin{scope}[start chain=going right, yshift=-7cm]
            \node[block] (Q) at (0,0) {China Stock Market Performance};
            \node[process] (R) {Resilience in China’s stock market bolsters global investor sentiment.};
            \node[block] (S) {Global Investor Sentiment};
            \node[process] (T) {Negative investor sentiment [...] leads to lower oil prices.};
        \end{scope}
    
        \begin{scope}[start chain=going left, yshift=-8.75cm]
            \node[block] (U) at (0.815\linewidth,0) {Crude Oil Prices};
            \node[process] (V) {Rising crude oil prices encourage increased drilling activity [...].};
            \node[block] (W) {US Oil Production};
            \node[process] (X) {US oil production contributes to [...] energy independence [...].};
        \end{scope}
    
        \begin{scope}[start chain=going right, yshift=-10.5cm]
            \node[block] (Y) at (0,0) {US Energy Independence};
            \node[process] (Z) {As the U.S. becomes energy independent, [...] threats to seize Syrian oil.};
            \node[block] (A1) {US Threat on Syrian Oil};
            \node[process] (B1) {Threat [...] amplifies overall geopolitical tensions.};
        \end{scope}
    
        \begin{scope}[start chain=going left, yshift=-12.25cm]
            \node[block] (C1) at (0.815\linewidth,0) {Geopolitical Tensions};
            \node[process] (D1) {Geopolitical tensions can lead to political destabilization and civil unrest in the Middle East.};
            \node[block] (E1) {Middle East Unrest};
            \node[process] (F1) {Geopolitical tensions [...] contributed to an initial spike in Canadian gas prices.};
        \end{scope}
    
        \begin{scope}[start chain=going right, yshift=-14cm]
            \node[block] (G1) at (0,0) {Canadian Gas Prices};
            \node[process] (H1) {Rise in Canadian gas prices contributes to a higher overall inflation rate in Canada.};
            \node[block] (I1) {Canada Inflation Rate};
        \end{scope}
    
        \draw[->,causal arrow] (A) -- (B);
        \draw[->,causal arrow] (B) -- (C);
        \draw[->,causal arrow] (C) -- (D);
        \draw[->,causal arrow] (D) -- (E);
        \draw[->,causal arrow] (E) -- (F);
        \draw[->,causal arrow] (F) -- (G);
        \draw[->,causal arrow] (G) -- (H);
        \draw[->,causal arrow] (H) -- (I);
        \draw[->,causal arrow] (I) -- (J);
        \draw[->,causal arrow] (J) -- (K);
        \draw[->,causal arrow] (K) -- (L);
        \draw[->,causal arrow] (L) -- (M);
        \draw[->,causal arrow] (M) -- (N);
        \draw[->,causal arrow] (N) -- (O);
        \draw[->,causal arrow] (O) -- (P);
        \draw[->,causal arrow] (P) -- (Q);
        \draw[->,causal arrow] (Q) -- (R);
        \draw[->,causal arrow] (R) -- (S);
        \draw[->,causal arrow] (S) -- (T);
        \draw[->,causal arrow] (T) -- (U);
        \draw[->,causal arrow] (U) -- (V);
        \draw[->,causal arrow] (V) -- (W);
        \draw[->,causal arrow] (W) -- (X);
        \draw[->,causal arrow] (X) -- (Y);
        \draw[->,causal arrow] (Y) -- (Z);
        \draw[->,causal arrow] (Z) -- (A1);
        \draw[->,causal arrow] (A1) -- (B1);
        \draw[->,causal arrow] (B1) -- (C1);
        \draw[->,causal arrow] (C1) -- (D1);
        \draw[->,causal arrow] (D1) -- (E1);
        \draw[->,causal arrow] (E1) -- (F1);
        \draw[->,causal arrow] (F1) -- (G1);
        \draw[->,causal arrow] (G1) -- (H1);
        \draw[->,causal arrow] (H1) -- (I1);
    \end{tikzpicture}
    \caption{Illustration of the longest direct causal path in the CausalWorld graph. It contains 18 nodes, represented in \textcolor{blue!60}{blue} boxes. Arrows represent causal dependencies. The description of the dependency is shown in dashed \textcolor{red!40}{red} boxes. Note that all nodes except for the root can have additional causal parents not shown in the chain and that the strength and function related to the causal relationships are not shown. }
    \label{fig:longest-causal-chain}
\end{figure}
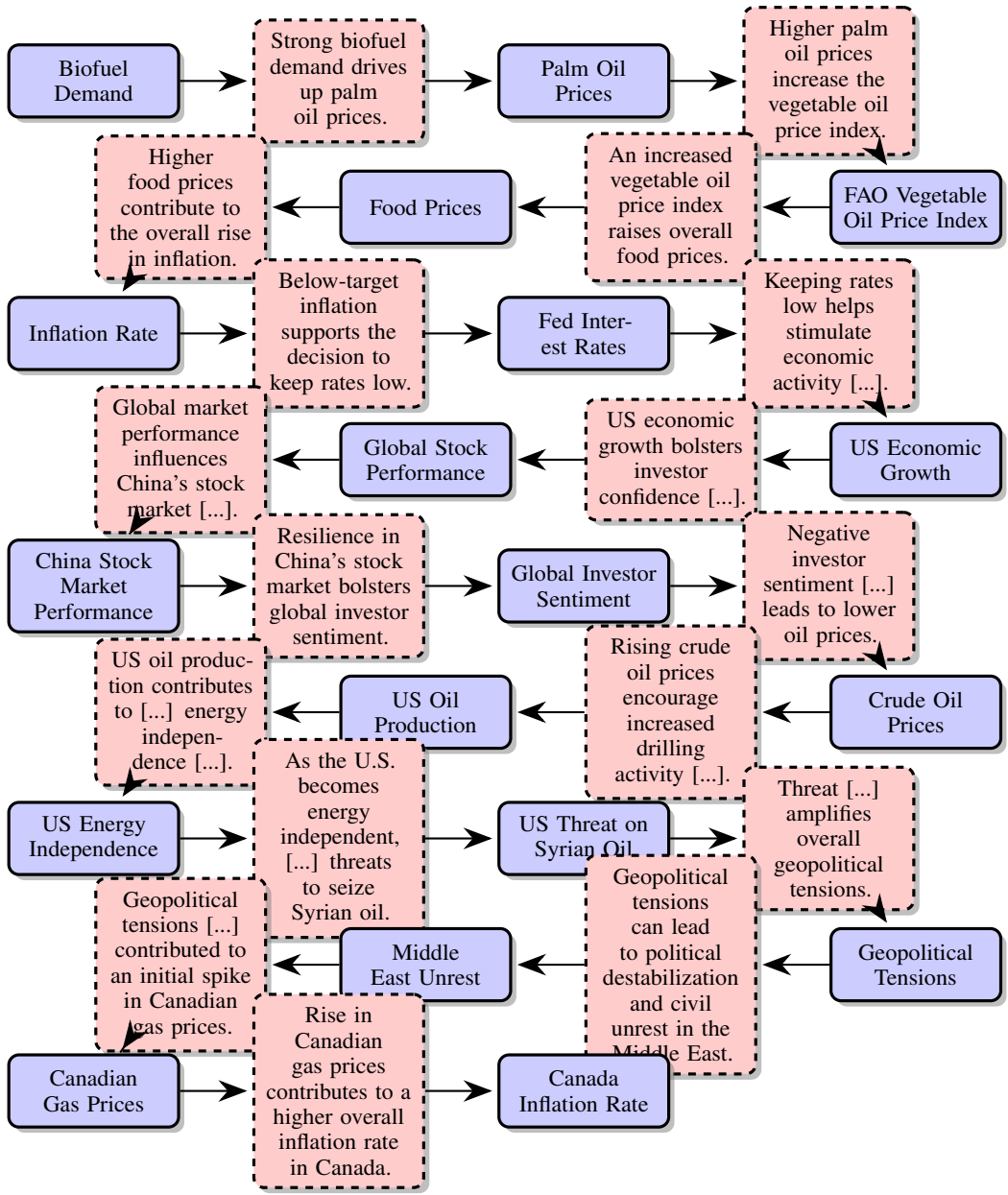

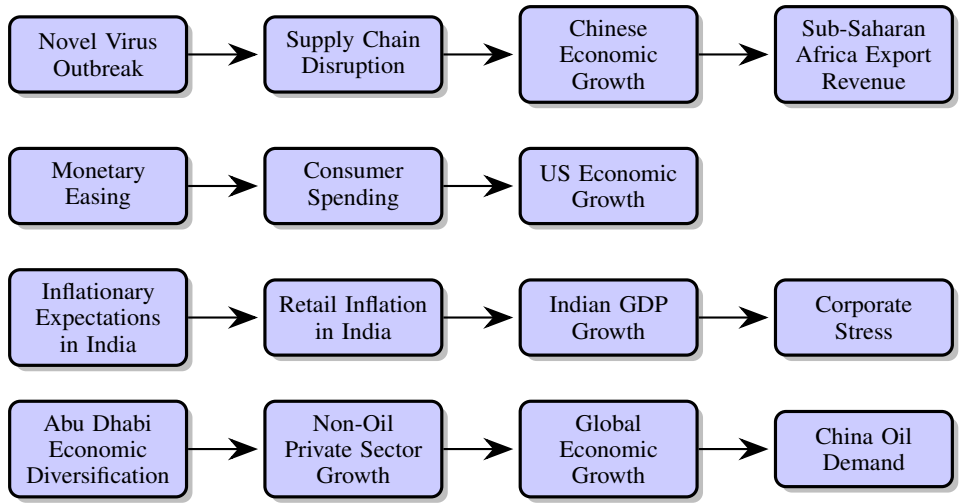
\begin{figure}
    \centering
    \begin{tikzpicture}[
        block/.style={
        rectangle, rounded corners,
        draw=black, very thick,
        font=\small,
        fill=blue!20,
        drop shadow, 
        minimum width=2cm, minimum height=1cm,
        align=center,
        inner sep=5pt,
        text width=2cm,
        },
        process/.style={
        ellipse,
        draw=black, very thick,
        font=\small,
        fill=red!20,
        drop shadow, 
        minimum width=2cm, minimum height=1cm,
        align=center,
        inner sep=5pt,
        text width=2cm,
        },
        causal arrow/.style={
        -{Stealth[length=4mm, width=3mm]},
        thick,
        shorten >=2pt
        },
        every node/.append style={on chain, node distance=1.5cm}
        ]
        \begin{scope}[start chain=going right]
            \node[block] (A) {Novel Virus Outbreak};
            \node[block] (B) {Supply Chain Disruption};
            \node[block] (C) {Chinese Economic Growth};
            \node[block] (D) {Sub-Saharan Africa Export Revenue};
        \end{scope}
    
        \begin{scope}[start chain=going right, yshift=-1.75cm]
            \node[block] (E) {Monetary Easing};
            \node[block] (F) {Consumer Spending};
            \node[block] (G) {US Economic Growth};
        \end{scope}
    
        \begin{scope}[start chain=going right, yshift=-3.5cm]
            \node[block] (H) {Inflationary Expectations in India};
            \node[block] (I) {Retail Inflation in India};
            \node[block] (J) {Indian GDP Growth};
            \node[block] (K) {Corporate Stress};
        \end{scope}
    
        \begin{scope}[start chain=going right, yshift=-5.25cm]
            \node[block] (L) {Abu Dhabi Economic Diversification};
            \node[block] (M) {Non-Oil Private Sector Growth};
            \node[block] (N) {Global Economic Growth};
            \node[block] (O) {China Oil Demand};
        \end{scope}
    
        \draw[->,causal arrow] (A) -- (B);
        \draw[->,causal arrow] (B) -- (C);
        \draw[->,causal arrow] (C) -- (D);
        
        \draw[->,causal arrow] (E) -- (F);
        \draw[->,causal arrow] (F) -- (G);
        
        \draw[->,causal arrow] (H) -- (I);
        \draw[->,causal arrow] (I) -- (J);
        \draw[->,causal arrow] (J) -- (K);
        
        \draw[->,causal arrow] (L) -- (M);
        \draw[->,causal arrow] (M) -- (N);
        \draw[->,causal arrow] (N) -- (O);
    \end{tikzpicture}
    \caption{Additional direct causal paths in CausalWorld graph. }
    \label{fig:causal-chain-ex2}
\end{figure}

Figure \ref{fig:longest-causal-chain} shows the longest causal chain that can be extracted from CausalWorld. Additional simplified chains are shown in Figure \ref{fig:causal-chain-ex2}.

\section{Details on the CausalWorld Counterfactual Reasoning Dataset}
\label{app:dataset_details}

\begin{figure}
    \centering
    \begin{tikzpicture}
    \begin{axis}[
        ybar,
        ymax=80,
        bar width=8pt,
        width=0.8\linewidth,
        height=5cm,
        xlabel={Number of node in query graph},
        ylabel={Occurences},
        symbolic x coords={2,3,4,5,6,7,8,9,10,11,12,13,14,15,16,17,18,19,23},
        xtick=data,
        xticklabel style={rotate=45, anchor=east},
        nodes near coords,
        enlarge x limits=0.1,
        every axis/.append style={font=\scriptsize}
        ]
    \addplot[fill=blue!60,draw=black] coordinates {
    (5, 28) (7, 29) (13, 16) (2, 60) (3, 41) (8, 20) (6, 12) (10, 27) (9, 36) (16, 8) (17, 4) (12, 55) (14, 18) (15, 2) (11, 11) (4, 28) (18, 5)
    };
    \end{axis}
    \end{tikzpicture}
    \caption{Distribution of the number of nodes in the query graphs of the CausalWorld-CR dataset. }
    \label{fig:dataset-graph-size}
\end{figure}
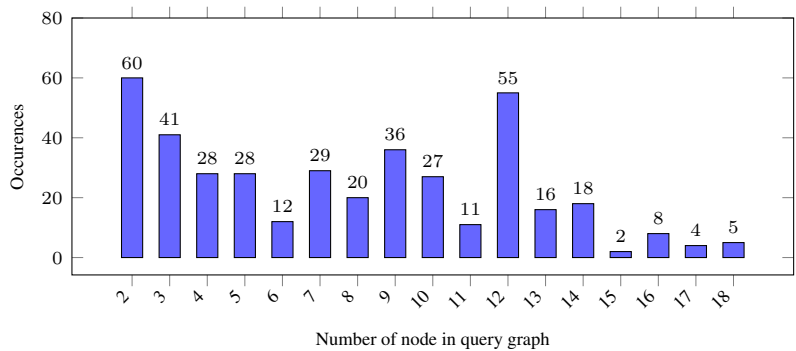

The CausalWorld network allows matching counterfactuals and building samples for causal queries. However, as shown in Figure \ref{fig:causal-world-components}, the degree distribution in the graph is imbalanced, resulting in an over-representation of some nodes over others. We balance the dataset to mitigate this issue. In addition, in the case of long dependencies, some queries may present many causal paths between the observations and the target variable. We remove queries with too many possible causal paths (i.e. $\geq 50$) to allow the problem to remain tractable. Figure \ref{fig:dataset-distrib} shows the distribution of query types and target domains in the CausalWorld-CR dataset. Figure \ref{fig:dataset-graph-size} shows the number of nodes per query graph.

\begin{figure}
    \centering
    \begin{subfigure}{0.42\linewidth}
        \begin{tikzpicture}
            \tikzstyle{every node}=[font=\scriptsize]
            \pie[
                    radius=1.4, 
                    text=legend, 
                    color={LimeGreen!60,Cyan!60},
                    font=\footnotesize
                ] 
            {50/Counterfactuals, 50/Observations} 
        \end{tikzpicture}
        \caption{Distribution of observation and counterfactual queries. }
    \end{subfigure}
    \hfill
    \begin{subfigure}{0.56\linewidth}
        \begin{tikzpicture}
            \tikzstyle{every node}=[font=\scriptsize]
            \pie[
                radius=2,
            	color = {
            		LimeGreen!60,
            		LimeGreen!60,
            		LimeGreen!60,
            		LimeGreen!60,
                    Cyan!60,
                    Cyan!60,
                    Cyan!60,
                    Cyan!60,
                    Cyan!60
            	}, 
            	explode = 0.1,
                font=\footnotesize
            ]
            {
                25.75/Crude Oil Prices,
            	10.75/Indian Rupee Exchange Rate,
            	8.25/Gold Prices,
                5.25/Other,
                3.25/Crude Oil Prices,
            	3.0/US Oil Production,
            	2.25/Asian Shares Performance,
            	2.25/US Oil and Gas Rig Count,
                39.25/Other
            }
        \end{tikzpicture}
        \caption{Distribution of target variables in the observation and counterfactual splits. }
    \end{subfigure}
    \caption{Distribution of the query types and target domains in the CausalWorld-CR dataset. Counterfactual queries are more concentrated around specific domains than observations because they require to match several worlds for building the query. }
    \label{fig:dataset-distrib}
\end{figure}
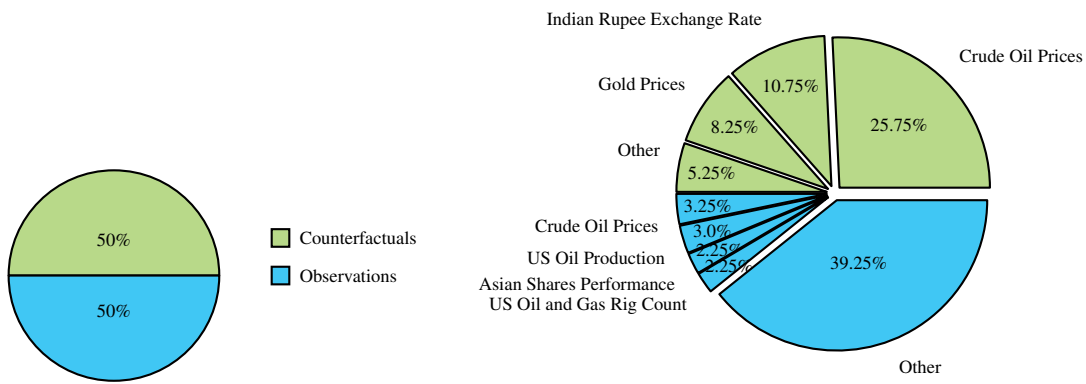

\section{Implementation Details}
\label{app:exp_details}

In this section, we detail the implementation details used for our experiments.

For the causal extraction pipeline, we use OpenAI o3-mini-2025-01-31 \citep{openai2023o3mini} as our base LLM. Specific details for NxGraphRAG are given in Section \ref{app:rag_details}. We use the Smolagents library\footnote{https://huggingface.co/docs/smolagents/} as the interface with the LLM agents.
We perform calls to o3-mini-2025-01-31 and gpt-4.1-2025-04-14 \citep{openai2025gpt41} using the LiteLLM interface with default hyperparameters. The order of magnitude for the total cost of the OpenAI API calls is $\sim \text{USD~} 100$. We run meta-llama/Llama-3.1-8B-Instruct \citep{meta2024llama31} locally using the Transformers interface. Local experiments are run on eight NVIDIA A100-SXM4-80GB GPUs.
The data analysis performed in Sections \ref{sec:causal_world} and \ref{sec:counterfactual-inference} involves semantic embedding. We perform the embedding using the all-mpnet-base-v2 \citep{all-mpnet-base-v2} Transformer model. We use the SentenceTransformers library \footnote{https://www.sbert.net/}. 
Data analysis experiments are performed on a single laptop with a 3.20 GHz AMD Ryzen 7 5800H CPU, 16GB RAM and a NVIDIA GeForce RTX 3070 Laptop GPU.

\section{Additional Results on CausalWorld-CR}
\label{app:additional_results}

\begin{figure}
    \begin{subfigure}{0.48\linewidth}
        \begin{tikzpicture}
            \begin{axis}[
                ybar,
                bar width=8pt,
                ymin=0,
                ymax=1.1,
                ylabel={Accuracy},
                symbolic x coords={o3-mini,GPT-4.1,LLaMA-3.1-8B},
                xtick=data,
                width=\linewidth,
                height=5cm,
                legend style={at={(1.0,0.96)},anchor=south east,legend columns=-1},
                enlarge x limits=0.25,
                xticklabel style={rotate=45, anchor=east},
                every axis/.append style={font=\scriptsize}
            ]
            \addplot+[fill=blue!60,draw=black] coordinates {(o3-mini,1) (GPT-4.1,1) (LLaMA-3.1-8B,0.82)};
            \addplot+[fill=orange!75!yellow!80,draw=black] coordinates {(o3-mini,0.76) (GPT-4.1,0.94) (LLaMA-3.1-8B,0.0)};
            \legend{CTG-Reason,CausalCoT}
            \end{axis}
            \node (s) at (4.42, 0.15) {*};
        \end{tikzpicture}
        \caption{Accuracy for boolean queries. }
        \label{fig:obs-bool-res}
    \end{subfigure}
    \hfill
    \begin{subfigure}{0.48\linewidth}
        \begin{tikzpicture}
            \begin{axis}[
                ybar,
                bar width=8pt,
                ymin=0,
                ymax=1.1,
                ylabel={Accuracy},
                symbolic x coords={o3-mini,GPT-4.1,LLaMA-3.1-8B},
                xtick=data,
                width=\linewidth,
                height=5cm,
                legend style={at={(1.0,0.96)},anchor=south east,legend columns=-1},
                enlarge x limits=0.25,
                xticklabel style={rotate=45, anchor=east},
                every axis/.append style={font=\scriptsize}
            ]
            \addplot+[fill=blue!60,draw=black] coordinates {(o3-mini,0.58) (GPT-4.1,0.51) (LLaMA-3.1-8B,0.52)};
            \addplot+[fill=orange!75!yellow!80,draw=black] coordinates {(o3-mini,0.53) (GPT-4.1,0.52) (LLaMA-3.1-8B,0.0)};
            \legend{CTG-Reason,CausalCoT}
            \end{axis}
            \node (s) at (4.42, 0.15) {*};
        \end{tikzpicture}
        \caption{Accuracy for trend queries. }
        \label{fig:obs-trend-res}
    \end{subfigure}
    \caption{Results on the \textit{observational} set for the boolean and trend queries of CausalWorld-CR. Results are shown for o3-mini, GPT-4o and LLaMA-3.1-8B, using CTG-Reason and CausalCoT. (*) The majority of queries with LLaMA-3.1-8B-CausalCoT returned with a timeout. }
    \label{fig:observation-results}
\end{figure}
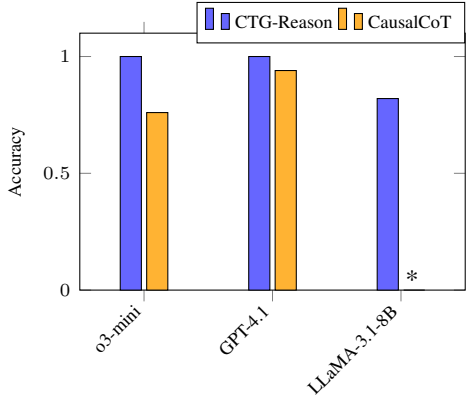
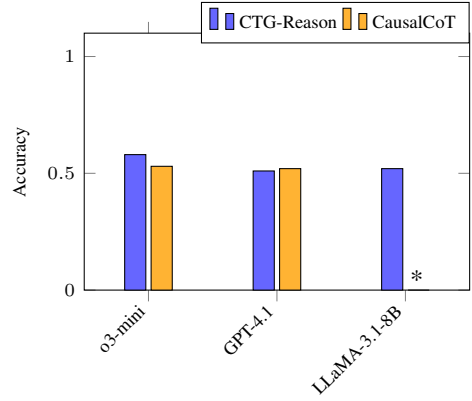

\begin{figure}
    \begin{subfigure}{0.48\linewidth}
        \includegraphics[width=\linewidth]{data/counterfactual-semantic-similarity.pdf}
        \caption{Counterfactual set. }
        \label{fig:ctf-text-semantic-sim}
    \end{subfigure}
    \hfill\begin{subfigure}{0.48\linewidth}
        \includegraphics[width=\linewidth]{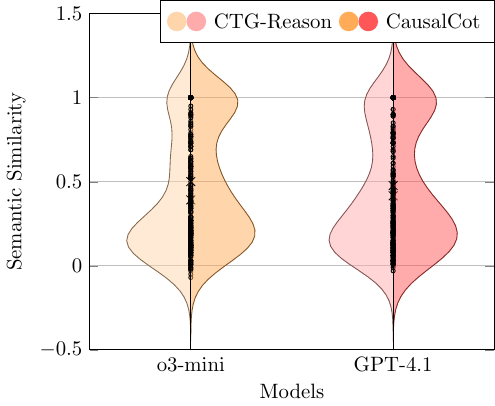}
        \caption{Observation set. }
        \label{fig:obs-text-semantic-sim}
    \end{subfigure}
    \caption{Violin plots of the semantic (cosine) similarity between ground truth and model answers. Results are shown for o3-mini and GPT-4.1. }
    \label{fig:semantic-sim}
\end{figure}

\begin{figure}
    \begin{subfigure}{0.48\linewidth}
        \includegraphics[width=\linewidth]{data/counterfactual-numerical-pct-error.pdf}
        \caption{Observation set. }
        \label{fig:ctf-numeric-err}
    \end{subfigure}
    \hfill
    \begin{subfigure}{0.48\linewidth}
        \includegraphics[width=\linewidth]{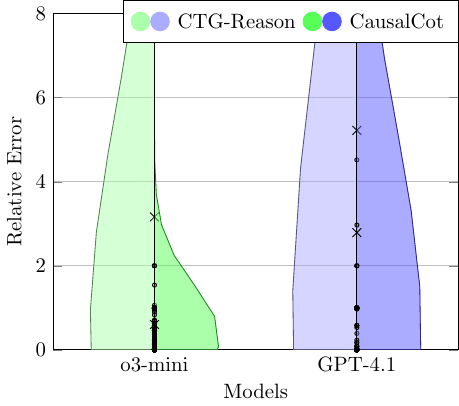}
        \caption{Observation set. }
        \label{fig:obs-numeric-err}
    \end{subfigure}
    \caption{Violin plots of the relative error numerical between ground truth and prediction. The results are shown for o3-mini and GPT-4.1 for samples where both ground truth and predicted values are numerical. The plots show the distribution of the error in \% of the ground truth value. }
    \label{fig:numeric-err-supp}
\end{figure}

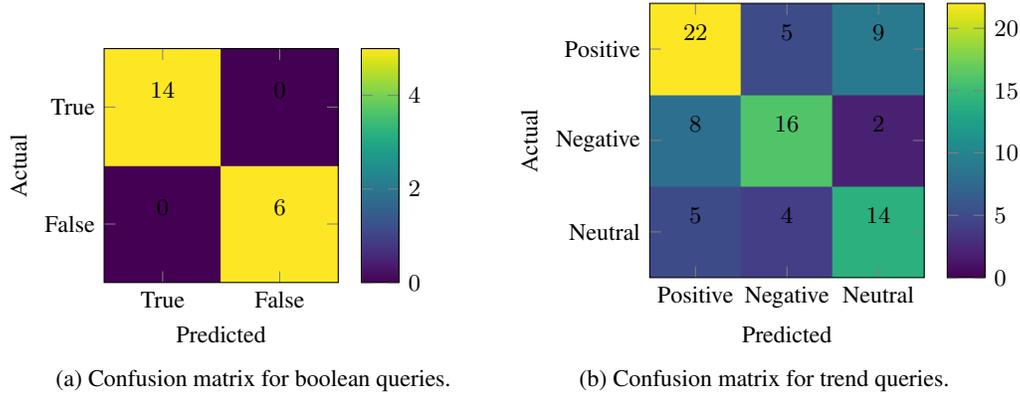
\begin{figure}
    \begin{subfigure}{0.48\linewidth}
        \begin{tikzpicture}
            \begin{axis}[
            width=0.7\linewidth,
            height=0.7\linewidth,
            xlabel={Predicted},
            ylabel={Actual},
            xtick={1,2},
            ytick={1,2},
            xticklabels={True,False},
            yticklabels={False,True},
            xmin=0.5, xmax=2.5,         
            ymin=0.5, ymax=2.5,
            axis equal image,         
            axis on top,
            colormap/viridis,
            colorbar,
            point meta min=0,
            point meta max=5,           
            mesh/cols=2,               
            every axis/.append style={font=\footnotesize},
            colorbar style={font=\footnotesize}
            ]
            
            \addplot [
              matrix plot*,            
              point meta=explicit,      
              nodes near coords,        
              every node near coord/.append style={font=\footnotesize\bfseries, color=black},
            ] table [meta index=2] {
              x y value
              1 1 0
              2 1 6
              1 2 14
              2 2 0
            };
            \end{axis}
        \end{tikzpicture}
        \caption{Confusion matrix for boolean queries. }
    \end{subfigure}
    \begin{subfigure}{0.48\linewidth}
        \begin{tikzpicture}
            \begin{axis}[
            width=0.78\linewidth,
            height=0.78\linewidth,
            xlabel={Predicted},
            ylabel={Actual},
            xtick={1,2,3},
            ytick={1,2,3},
            xticklabels={Positive,Negative,Neutral},
            yticklabels={Neutral,Negative,Positive},
            xmin=0.5, xmax=3.5,         
            ymin=0.5, ymax=3.5,
            axis equal image,         
            axis on top,
            colormap/viridis,
            colorbar,
            point meta min=0,
            point meta max=22,           
            mesh/cols=3,               
            every axis/.append style={font=\footnotesize},
            colorbar style={font=\footnotesize}
            ]
            
            \addplot [
              matrix plot*,            
              point meta=explicit,      
              nodes near coords,        
              every node near coord/.append style={font=\footnotesize\bfseries, color=black},
            ] table [meta index=2] {
              x y value
              1 1 5
              1 2 8
              1 3 22
              2 1 4
              2 2 16
              2 3 5
              3 1 14
              3 2 2
              3 3 9
            };
            \end{axis}
        \end{tikzpicture}
        \caption{Confusion matrix for trend queries. }
    \end{subfigure}
    \caption{Confusion matrices for boolean and trend queries of CausalWorld-CR for o3-mini using CTG-Reason. }
    \label{fig:confusion-matrices}
\end{figure}

\begin{figure}
    \centering
    \begin{subfigure}{0.48\linewidth}
        \includegraphics[width=\linewidth]{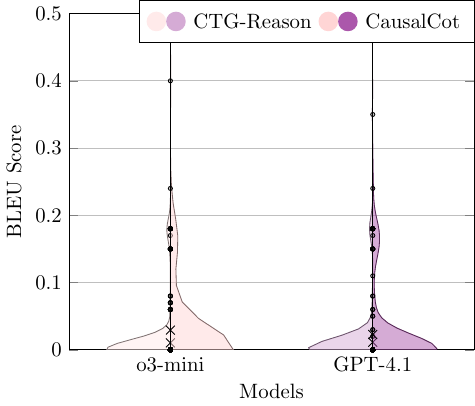}
        \caption{Counterfactual set.}
    \end{subfigure}
    \hfill
    \begin{subfigure}{0.48\linewidth}
        \includegraphics[width=\linewidth]{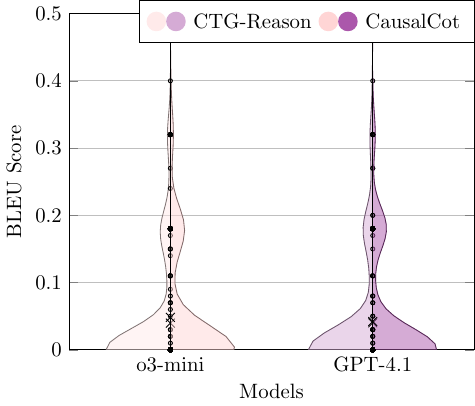}
        \caption{Observation set.}
    \end{subfigure}
    \caption{Violin plots of the BLEU score for o3-mini and GPT-4.1 using CTG-Reason on counterfactual and observational query sets. A higher score indicates that the model prediction contains n-grams similar to the ones appearing in the ground truth answer. }
    \label{fig:bleu-score}
\end{figure}

Figures \ref{fig:counterfactual-results} and \ref{fig:observation-results} show the detailed results on the counterfactual and observation queries of CausalWorld-CR. The main results can be found in Figure \ref{fig:main-results}. Figure \ref{fig:confusion-matrices} further shows the confusion matrices for the boolean and trend queries while Figures \ref{fig:semantic-sim} and \ref{fig:numeric-err-supp} show the distributions of the cosine similarity and relative error between the ground truth and the predictions, respectively.
Figure \ref{fig:bleu-score} shows the BLEU scores \citep{DBLP:conf/acl/PapineniRWZ02} for o3-mini and GPT-4.1 on the counterfactual and observational sets. Higher scores indicates that model responses contains n-grams similar to the ones appearing in ground truth answers. While BLEU is suited for translation, in this settings, it indicates how close the answer formulation is compared to the original. A low score does not indicate that the answer is incorrect but that the grammatical elements that are used differ. As expected, score are indeed low but observational answers obtain slightly higher scores.

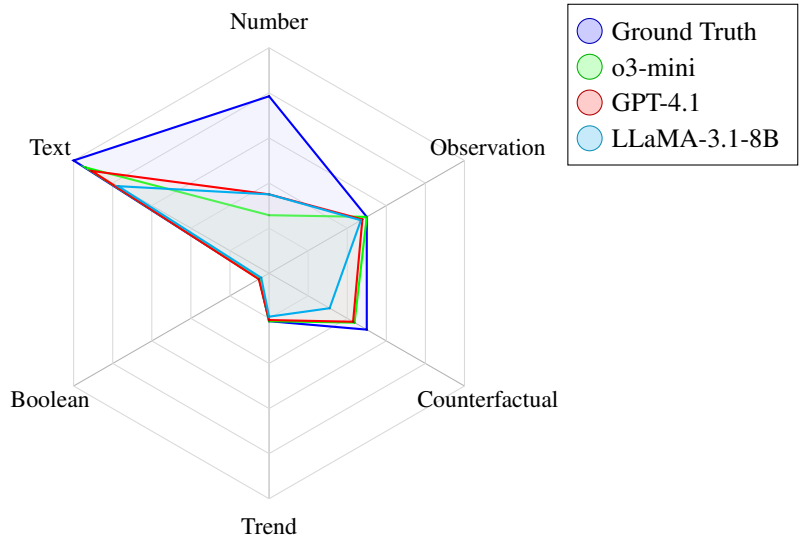
\begin{figure}
    \centering
    \begin{tikzpicture}
        \def\n{6}                 
        \def\R{3}                
        \def\levels{5}           
        \def\labels{Number,Text,Boolean,Trend,Counterfactual,Observation} 
        \def\dataA{0.2575, 0.9375, 0.05, 0.2125, 0.4375, 0.5} 
        \def\dataB{0.35, 0.9075, 0.0525, 0.2075, 0.43, 0.4775} 
        \def\dataC{0.35, 0.7725, 0.04, 0.1925, 0.31, 0.47} 
        \def\gtdata{0.785, 1.0, 0.05, 0.2125, 0.5, 0.5} 
        
        \kiviatgrid{(0,0)}{\R}{\n}{\levels}
        
        \foreach \labeltext [count=\i,evaluate={\angle=90+360/\n*(\i-1);}] in \labels {
            \node[font=\footnotesize,align=center] at ({\angle}:\R*1.12) {\labeltext};
        }
        
        \foreach \val [count=\i,evaluate={\angle = 90+360/\n*(\i-1);}] in \gtdata {
            \coordinate (G\i) at ({\angle}:{\R*\val});
        }
        
        \foreach \val [count=\i,evaluate={\angle = 90+360/\n*(\i-1);}] in \dataA {
            \coordinate (D\i) at ({\angle}:{\R*\val});
        }
        
        \foreach \val [count=\i,evaluate={\angle = 90+360/\n*(\i-1);}] in \dataB {
            \coordinate (E\i) at ({\angle}:{\R*\val});
        }
        
        \foreach \val [count=\i,evaluate={\angle = 90+360/\n*(\i-1);}] in \dataC {
            \coordinate (F\i) at ({\angle}:{\R*\val});
        }
          
        \filldraw[thick,blue,fill=blue!20,fill opacity=0.2]
          (G1) \foreach \i in {2,...,\n} { -- (G\i) } -- cycle;
        
        \foreach \i in {1,...,\n}
          \fill[blue!70!black] (G\i) circle (0.015);
        
        \filldraw[thick,green,fill=green!20,fill opacity=0.2]
          (D1) \foreach \i in {2,...,\n} { -- (D\i) } -- cycle;
        
        \foreach \i in {1,...,\n}
          \fill[green!70!black] (D\i) circle (0.015);
        
        \filldraw[thick,red,fill=red!20,fill opacity=0.2]
          (E1) \foreach \i in {2,...,\n} { -- (E\i) } -- cycle;
        
        \foreach \i in {1,...,\n}
          \fill[red!70!black] (E\i) circle (0.015);
        
        \filldraw[thick,cyan,fill=cyan!20,fill opacity=0.2]
          (E1) \foreach \i in {2,...,\n} { -- (F\i) } -- cycle;
        
        \foreach \i in {1,...,\n}
          \fill[cyan!70!black] (F\i) circle (0.015);

        \matrix [draw,below left,anchor=north west] at (current bounding box.north east) {
          \node [shape=circle,fill=blue!20,draw=blue!70!black,label=right:Ground Truth] {}; \\
          \node [shape=circle,fill=green!20,draw=green!70!black,label=right:o3-mini] {}; \\
          \node [shape=circle,fill=red!20,draw=red!70!black,label=right:GPT-4.1] {}; \\
          \node [shape=circle,fill=cyan!20,draw=cyan!70!black,label=right:LLaMA-3.1-8B] {}; \\
        };
    \end{tikzpicture}
    \caption{Distribution of query types and answer types from CTG-Reason. Types correspond to a category a sample can be evaluated in. Types are: \textit{number} (answer is a number), \textit{text} (answer is textual, matches all samples), \textit{boolean} (true/false), \textit{trend} (the qualitative assessment of a trend: increasing, decreasing, stable). \textit{Observation} and \textit{counterfactual} correspond to thequery types as defined i Section \ref{sec:counterfactual-inference}. A sample can cumulate multiple types and the model answer may not correspond to the same type as the ground truth. For instance, while the ground-truth has a numerical value, the model may provide a qualitative answer. }
    \label{fig:label-distrib}
\end{figure}

\begin{figure}
    \begin{subfigure}{0.48\linewidth}
        \begin{tikzpicture}
            \begin{axis}[
                ybar,
                bar width=8pt,
                ymin=0,
                ymax=1.1,
                ylabel={Accuracy},
                symbolic x coords={o3-mini,GPT-4.1,LLaMA-3.1-8B},
                xtick=data,
                width=\linewidth,
                height=5cm,
                legend style={at={(1.0,0.96)},anchor=south east,legend columns=-1},
                enlarge x limits=0.25,
                xticklabel style={rotate=45, anchor=east},
                every axis/.append style={font=\scriptsize}
            ]
            \addplot+[fill=blue!60,draw=black] coordinates {(o3-mini,1) (GPT-4.1,1) (LLaMA-3.1-8B,0.8)};
            \addplot+[fill=orange!75!yellow!80,draw=black]  coordinates {(o3-mini,1) (GPT-4.1,1) (LLaMA-3.1-8B,0.0)};
            \legend{CTG-Reason,CausalCoT}
            \end{axis}
            \node (s) at (4.42, 0.15) {*};
        \end{tikzpicture}
        \caption{Accuracy for boolean queries. }
        \label{fig:ctf-bool-res}
    \end{subfigure}
    \hfill
    \begin{subfigure}{0.48\linewidth}
        \begin{tikzpicture}
            \begin{axis}[
                ybar,
                bar width=8pt,
                ymin=0,
                ymax=1.1,
                ylabel={Accuracy},
                symbolic x coords={o3-mini,GPT-4.1,LLaMA-3.1-8B},
                xtick=data,
                width=\linewidth,
                height=5cm,
                legend style={at={(1.0,0.96)},anchor=south east,legend columns=-1},
                enlarge x limits=0.25,
                xticklabel style={rotate=45, anchor=east},
                every axis/.append style={font=\scriptsize}
            ]
            \addplot+[fill=blue!60,draw=black] coordinates {(o3-mini,0.82) (GPT-4.1,0.9) (LLaMA-3.1-8B,0.55)};
            \addplot+[fill=orange!75!yellow!80,draw=black] coordinates {(o3-mini,0.9) (GPT-4.1,0.7) (LLaMA-3.1-8B,0.0)};
            \legend{CTG-Reason,CausalCoT}
            \end{axis}
            \node (s) at (4.42, 0.15) {*};
        \end{tikzpicture}
        \caption{Accuracy for trend queries. }
        \label{fig:ctf-trend-res}
    \end{subfigure}
    \caption{Results on the \textit{counterfactual} set for the boolean and trend queries of CausalWorld-CR. Results are shown for o3-mini, GPT-4o and LLaMA-3.1-8B, using CTG-Reason and CausalCoT. (*) The majority of queries with LLaMA-3.1-8B-CausalCoT returned with a timeout. }
    \label{fig:counterfactual-results}
\end{figure}
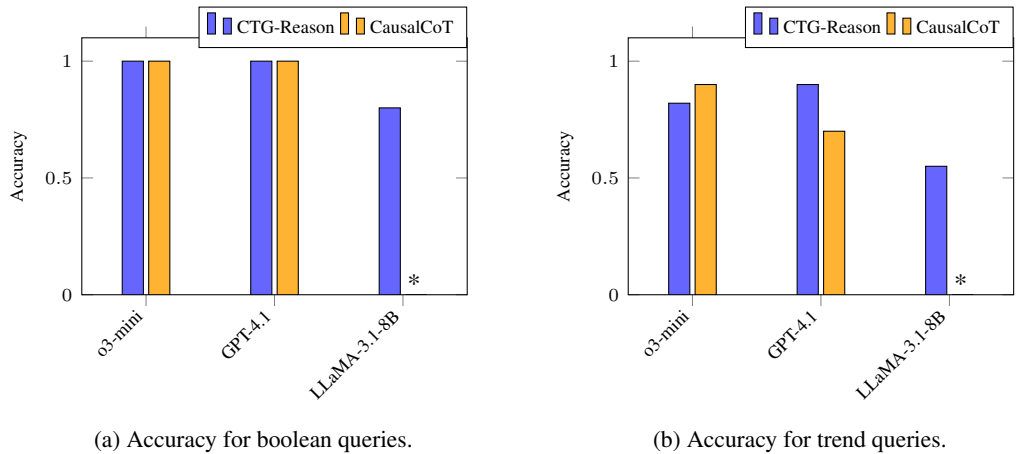

\section{Agent Prompts}
\label{app:prompts}

This section describes the system and user prompts used for the LLM agents in the extraction and inference systems, as well as the syntax for the causal variables and relationships in the causal network.

\subsection{Causal Information Syntax}

The syntax for the causal variables and relationships is provided in the LLMs' prompts as shown in the next sections. Additionally, it is also enforced using Pydantic.

\subsubsection{Causal Variables}

Causal variables are defined in the causal network as a dictionary of attributes. They are described to the LLM agents as follows:

\begin{lstlisting}
{
    "name": <string>, # The name of the variable
    "description": <string>, # The description of the variable
    "type": <string>, # The type of the variable (boolean, integer, float, string, etc.)
    "values": <List[str]>, # The set or range of possible values for the variable ([1, 2, 3], 'range(0,10)', ['low', 'medium', 'high'], 'True/False', 'natural numbers', etc.)
    "causal_effect": <string>, # The inferred causal effect of the variable
    "supporting_text_snippets": <Optional[List[str]]>, # The supporting text snippets in which the variable is mentioned
    "current_value": <Optional[string]>, # The observed current value of the variable
    "contextual_information": <Optional[string]>, # The contextual information associated with the current value of the variable
}
\end{lstlisting}

See Section \ref{sec:causal_extraction} in the main paper for more details.

\subsubsection{Causal Relationships}

Similarly to causal variables, causal relationships are defined in the causal network as a dictionary of attributes. They are described to the LLM agents as follows:

\begin{lstlisting}
{
    "cause": <string>, # The name of the cause variable
    "effect": <string>, # The name of the effect variable
    "description": <string>, # The description of the causal relationship between the variables
    "contextual_information": <Optional[string]>, # The contextual information associated with the causal relationship for the specific observed values of the variables
    "type": <string>, # The type of the causal relationship (direct, indirect, etc.)
    "strength": <Optional[string]>, # The strength of the causal relationship
    "confidence": <Optional[string]>, # The confidence level in the existence of the causal relationship
    "function": <Optional[Callable]>, # The function that describes the causal relationship, if available.
}
\end{lstlisting}

\subsection{Causal Extraction Agent Prompt}

We show below the start of the causal extraction agent system prompt. It describes to the agent the extraction task to be performed and how to solve it. It is built upon on the ReAct framework \citep{DBLP:conf/iclr/YaoZYDSN023} and the prompts provided by the Smolagents library \footnote{https://huggingface.co/docs/smolagents/}.

\begin{lstlisting}
  You are an expert assistant who can solve any task using code blobs. You specialize into causal extraction tasks. 
  You will be given a text snippet and an initial causal graph. Your task will consist of finding the causal variables described in the text, the causal relationships that link them and adding them to the causal graph.
  You will solve the task as best you can. To do so, you have been given access to a Python interpreter with the standard library and the networkx package. 
  You will also have access to an optional list of tools: these tools are basically Python functions which you can call with code.
  You will use your expert reading comprehension, commonsense reasoning and coding skills to tolve the problem.
  To solve the task, you must plan forward to proceed in a series of steps, in a cycle of 'Thought:', 'Code:', and 'Observation:' sequences.

  At each step, in the 'Thought:' sequence, you should first explain your reasoning towards solving the task and the tools that you want to use.
  Then in the 'Code:' sequence, you should write the code in simple Python. The code sequence must end with '<end_code>' sequence.
  During each intermediate step, you can use 'print()' to save whatever important information you will then need.
  These print outputs will then appear in the 'Observation:' field, which will be available as input for the next step.
  In the end, you have to return a final answer using the `final_answer` tool. The output provided to the `final_answer` tool should be the networkx causal graph.

  Each node should have the following dictionary of attributes:
  {{variable}}
  Some variables will have a value provided in the text, while others will be confounders that need to be estimated. Provide a current~value and contextual information whenever possible.

  The causal relationships should be represented as directed edges between the nodes. Each edge should have the following dictionary of attributes:
  {{causal_relationship}}

  Your plan should be as follows:
  1. Define the causal variables observed in the text. Use the variables provided when possible or create new ones when no variable matches.
  2. Define the confounders that are not observed in the text or for which a value is not given, and that affect one or several of the causal variables defined in step 1.
  3. Verify if the new variables have correspondance in the causal graph database. Use the `{{retrieval_tool_name}}` tool to assess if the variable is already in the database. 
  If it is, use it instead of creating a new one. It may have a different name in the database, the tool returns the top-k matching variables. 
  THIS IS A MANDATORY STEP. The variables provided in the input are only a subset of the variables in the database, you should always check if the variable already exist before creating new ones.
  For each variable, use the one matching the most or create a new one if none matches.
  4. Define the causal relationships between the variables, based on the text and common sense knowledge. Do not create causal relationships that already exist in the causal graph.
  5. Build the full causal graph as a networkx DiGraph object.
  Each step should be a separate 'Thought:', 'Code:', and 'Observation:' sequence.
  
  The code MUST be executed in two code blocks minimum. After step 3, use <end_code> to indicate the end of the code block and retrieve the output of the tool call. 
  Then, use the observation to execute steps 4 and 5 and complete the task. An empty list of variables in the user input DOES NOT MEAN that there are no variables in the causal graph database.
  Always check if the variable exists in the database before creating a new one.

[...]
\end{lstlisting}

The \texttt{variable} and \texttt{causal\_relationship} placeholders are replaced with the prompts described in the previous section.
Few-shots are also provided as part of the system prompt. Here is one example:

\begin{lstlisting}[mathescape=true]
[...]

Here are a few examples tasks to solve.
---
Task:
"""
Increased Air Pollution Leads to Higher Respiratory Issues During Summer Months

A recent study has shown that cities experiencing high levels of air pollution are seeing a rise in respiratory issues among their populations, particularly during the summer months. 
Experts attribute the increase to a combination of higher temperatures and pollution from industrial activities, which intensify during hotter months. 
Dr. Laura Simpson, a pulmonologist, noted that individuals with pre-existing conditions, such as asthma, are at a greater risk of suffering from exacerbated symptoms.

Retrieved nodes:
<empty>

Retrieved edges:
<empty>
"""

Thought: I will first define the list of causal variables extracted from the text. These include:
1. Air Pollution - representing high levels of air pollution in cities.
2. Temperature - higher temperatures during the summer months.
3. Industrial Pollution - arising from intensified industrial activities during hotter months.
4. Respiratory Issues - the rise in respiratory issues among the population.
5. Pre-existing Respiratory Conditions - representing vulnerable individuals (e.g., those with asthma) at greater risk.

I will verify if these variables exist in our causal graph database using the tool graph_retriever for each variable by calling it with a query that includes the variable name. Then I will build the networkx DiGraph with these nodes and add the appropriate directed edges based on cause-effect relationships:
- Air Pollution $\rightarrow$ Respiratory Issues ("Increased air pollution leads to higher respiratory issues.")
- Temperature $\rightarrow$ Industrial Pollution ("Hotter temperatures intensify industrial activities' pollution.")
- Industrial Pollution $\rightarrow$ Respiratory Issues ("Pollution from industrial activities contributes to respiratory issues.")
- Pre-existing Respiratory Conditions $\rightarrow$ Respiratory Issues ("Individuals with pre-existing conditions are more vulnerable to respiratory issues.")

I will now call graph_retriever to check for each variable.

Code:
```py
import networkx as nx

# Define initial variables extracted from the text
causal_variables = [
{
    "name": "Air Pollution",
    "description": "The level of air pollutants present in the city environment.",
    "type": "float",
    "values": "range(0,100)",
    "supporting_text_snippets": ["cities experiencing high levels of air pollution"],
    "current_value": None,
    "contextual_information": "High in some cities as noted in the study."
},
{
    "name": "Temperature",
    "description": "Ambient temperature, particularly during the summer months.",
    "type": "float",
    "values": "range(20,50)",
    "supporting_text_snippets": ["higher temperatures", "summer months"],
    "current_value": None,
    "contextual_information": "Elevated during summer."
},


(truncated)


]

# Query the causal graph database for each variable using graph_retriever
for variable in causal_variables:
query_text = f"{variable['name']}: {variable['description']}"
result = graph_retriever(query=query_text)
print(f"Query for variable '{variable['name']}':", result, '\n')

print("End of variable verification.")
```<end_code>
Observation:
Query for variable 'Air Pollution': Retrieved nodes:
<empty>

Retrieved edges:
<empty>

Query for variable 'Temperature': Retrieved nodes:
<empty>

Retrieved edges:
<empty>


(truncated)


End of variable verification.

Thought: None of our causal variables exist in the database, so I will now build the full networkx DiGraph using our defined variables (as nodes) and add the causal relationships (as directed edges). 
I will add the following edges:
- Air Pollution --> Respiratory Issues: "Increased air pollution leads to higher respiratory issues."
- Temperature --> Industrial Pollution: "Higher temperatures intensify industrial pollution from industrial activities."
- Industrial Pollution --> Respiratory Issues: "Pollution from industrial activities contributes to respiratory issues."
- Pre-existing Respiratory Conditions --> Respiratory Issues: "Individuals with pre-existing respiratory conditions are more vulnerable to respiratory issues."

I will now add these nodes and edges to the graph and return the final causal graph.

Code:
```py
# Build the full causal graph as a networkx DiGraph
G = nx.DiGraph()

# Add nodes for each variable and update their attributes separately
for variable in causal_variables:
var_name = variable["name"]
G.add_node(var_name)
G.nodes[var_name].update(variable)

# Define causal relationships as edges with descriptive attributes
causal_relationships = [
{
    "cause": "Air Pollution",
    "effect": "Respiratory Issues",
    "description": "Increased air pollution in cities leads to higher respiratory issues among populations.",
    "contextual_information": "Study shows a rise in respiratory issues in high pollution cities during summer.",
    "type": "direct",
    "strength": "high",
    "confidence": "high",
    "function": None
},
{
    "cause": "Temperature",
    "effect": "Industrial Pollution",
    "description": "Higher temperatures intensify industrial activities, leading to increased industrial pollution.",
    "contextual_information": "Pollution from industrial activities intensifies during hotter months.",
    "type": "direct",
    "strength": "moderate",
    "confidence": "high",
    "function": None
},


(truncated)


]

# Add edges from the causal_relationships list
for edge in causal_relationships:
cause = edge["cause"]
effect = edge["effect"]
G.add_edge(cause, effect)
G[cause][effect].update(edge) # /!\ This is how edge attributes should be added. They must be added after the edge is created. 

# Output the final graph details for verification
print("Causal Graph Nodes:")
print(list(G.nodes(data=True)))
print("\nCausal Graph Edges:")
print(list(G.edges(data=True)))

final_answer(G)
```<end_code>

[...]
\end{lstlisting}

Three examples are provided to the LLM. The rest of the prompt is identical to the default system prompt of Smolagents.

\subsection{Counterfactual Inference Agent Prompt}

We show below the start of the counterfactual inference agent system prompt, it is built similarly to the causal extraction agent.

\begin{lstlisting}
You are an expert assistant who can solve any task using code blobs. You specialize into causal inference tasks. 
You will be given causal variables and causal relationships. Your task will consist of computing the causal effect of a target causal variable given the values of its direct parents or children in the causal graph.
You will solve the task as best you can. To do so, you have been given access to a Python interpreter with the standard library. 
You will also have access to an optional list of tools: these tools are basically Python functions which you can call with code.
You will use your expert reading comprehension, commonsense reasoning and coding skills to tolve the problem.
To solve the task, you must plan forward to proceed in a series of steps, in a cycle of 'Thought:', 'Code:', and 'Observation:' sequences.

At each step, in the 'Thought:' sequence, you should first explain your reasoning towards solving the task and the tools that you want to use.
Then in the 'Code:' sequence, you should write the code in simple Python. The code sequence must end with '<end_code>' sequence.
During each intermediate step, you can use 'print()' to save whatever important information you will then need.
These print outputs will then appear in the 'Observation:' field, which will be available as input for the next step.
In the end, you have to return a final answer using the `final_answer` tool. The output provided to the `final_answer` tool should be the networkx causal graph.

The attributes of the target variable are provided as arguments with the name 'target_variable'. 
The parent variables attributes are provided as a list of dictionaries with the name 'parent_variables'. 
The children variables attributes are provided as a list of dictionaries with the name 'children_variables'. 
The descriptions of the causal relationships between the target variable and its parents are provided as a list of attribute dictionaries with the name 'causal_relationships'.
Return a dictionary with the updated attributes of the target variable. The fields to update are 'current_value', 'contextual_information' and 'causal_effect'
Use the information from the given target variable, parent variables, children variables and causal relationships to compute the causal effect of the target variable.
When a function is provided in the causal relationship, execute it with the Python interpreter.

Each causal variable will be represented with the following dictionary of attributes:
{{variable}}

The causal relationships will be represented with the following dictionary of attributes:
{{causal_relationship}}
\end{lstlisting}

Five few-shots are also provided as part of the system prompt. Here is one example:

\begin{lstlisting}
[...]

Here are a few examples tasks to solve.
---
Task:
"""
Compute the causal effect of the variables in the context of the oil price war 

You have been provided with these additional arguments, that you can access using the keys as variables in your python code:
{'children_variables': [], 'target_variable': {'name': 'oil_prices', 'description': 'Global price per barrel of oil; recently slumped to its lowest level since 2009', 'type': 'float', 'values': 'USD per barrel'}, 'causal_relationships': []}.
"""

Thought: We have a target variable "oil_prices" with description indicating that oil prices have slumped to the lowest level since 2009. 
There are no parent or children variables provided and no causal relationships specified. 
In this context, we need to update the target_variable dictionary by setting its "current_value", "contextual_information", and "causal_effect". 
Since the variable is operating in an environment described as an "oil price war", we can infer that the aggressive competition among oil producers is likely driving a strong negative causal impact on oil_prices. 

I'll now update the target variable accordingly:
- current_value: "lowest level since 2009"
- contextual_information: "In the context of an oil price war, aggressive market competition among oil-producing nations has driven global oil prices down significantly."
- causal_effect: "lowest level since 2009"

I'll now write the code with these updates and then use final_answer to output the updated target_variable.

Code:
```py
# Update target_variable with computed values
target_variable['current_value'] = "lowest level since 2009"
target_variable['contextual_information'] = ("In the context of an oil price war, aggressive market competition among oil-producing nations has driven global oil prices down significantly.")
target_variable['causal_effect'] = "lowest level since 2009"

final_answer(target_variable)
```<end_code>

[...]
\end{lstlisting}



\end{document}